\documentclass[journal]{IEEEtran}

\usepackage{cite}

\usepackage{xcolor}
\usepackage{multirow}
\usepackage{pbox}
\usepackage{amsmath}
\usepackage{amssymb}
\usepackage{hyperref}
\usepackage{url}
\usepackage{colortbl}
\usepackage{graphicx}
\usepackage{adjustbox}
\usepackage{subfig}
\usepackage{makecell}
\usepackage{float}
\usepackage{scalerel}
\usepackage{tikz}
\usepackage{float}
\usetikzlibrary{svg.path}

\definecolor{orcidlogocol}{HTML}{A6CE39}
\tikzset{
  orcidlogo/.pic={
    \fill[orcidlogocol] svg{M256,128c0,70.7-57.3,128-128,128C57.3,256,0,198.7,0,128C0,57.3,57.3,0,128,0C198.7,0,256,57.3,256,128z};
    \fill[white] svg{M86.3,186.2H70.9V79.1h15.4v48.4V186.2z}
                 svg{M108.9,79.1h41.6c39.6,0,57,28.3,57,53.6c0,27.5-21.5,53.6-56.8,53.6h-41.8V79.1z M124.3,172.4h24.5c34.9,0,42.9-26.5,42.9-39.7c0-21.5-13.7-39.7-43.7-39.7h-23.7V172.4z}
                 svg{M88.7,56.8c0,5.5-4.5,10.1-10.1,10.1c-5.6,0-10.1-4.6-10.1-10.1c0-5.6,4.5-10.1,10.1-10.1C84.2,46.7,88.7,51.3,88.7,56.8z};
  }
}

\newcommand\orcidicon[1]{\href{https://orcid.org/#1}{\mbox{\scalerel*{
\begin{tikzpicture}[yscale=-1,transform shape]
\pic{orcidlogo};
\end{tikzpicture}
}{|}}}}

\definecolor{lightgray}{rgb}{0.92, 0.92, 0.92}

\ifCLASSINFOpdf
\else
\fi
\begin{document}

\title{\vspace{-3mm} Graph Convolutional Networks based on Manifold Learning for Semi-Supervised Image Classification}

\author{Lucas Pascotti Valem \orcidicon{0000-0002-3833-9072},
    ~Daniel Carlos Guimarães Pedronette \orcidicon{0000-0002-2867-4838} \\ 
    and~Longin Jan Latecki \orcidicon{0000-0002-5102-8244}. %\IEEEmembership{Senior Member,~IEEE \vspace{-5mm}}% <-this % stops a space
\thanks{L. P. Valem and D. C. G. Pedronette are with the Department of Statistics,
Applied Math. and Computing, State University of São Paulo, Rio Claro, Brazil (e-mail: \{lucas.valem, daniel.pedronette\}@unesp.br).}% <-this % stops a space
\thanks{L. J. Latecki is with Department of Computer and Information Sciences, Temple University, Philadelphia, USA (e-mail: latecki@temple.edu).}% <-this % stops a space
}% <-this % stops a space
\markboth{Accepted version of paper published in Computer Vision and Image Understanding}%
{Valem\MakeLowercase{\textit{et al.}}}

\maketitle

\begin{abstract}
Due to a huge volume of information in many domains, the need for classification methods is imperious.
In spite of many advances, most of the approaches require a large amount of labeled data, which is often not available, due to costs and difficulties of manual labeling processes. 
In this scenario, unsupervised and semi-supervised approaches have been gaining increasing attention. 
The GCNs (Graph Convolutional Neural Networks) represent a promising solution since they encode the neighborhood information and have achieved state-of-the-art results on scenarios with limited labeled data.
However, since GCNs require graph-structured data, their use for semi-supervised image classification is still scarce in the literature.
In this work, we propose a novel approach, the Manifold-GCN, based on GCNs for semi-supervised image classification.
The main hypothesis of this paper is that the use of manifold learning to model the graph structure can further improve the GCN classification.
To the best of our knowledge, this is the first framework that allows the combination of GCNs with different types of manifold learning approaches for image classification. All manifold learning algorithms employed are completely unsupervised, which is especially useful for scenarios where the availability of labeled data is a concern.
A broad experimental evaluation was conducted considering 5 GCN models, 3 manifold learning approaches, 3 image datasets, and 5 deep features.
The results reveal that our approach presents better accuracy than traditional and recent state-of-the-art methods with very efficient run times for both training and testing.
\end{abstract}

%\begin{IEEEkeywords}
%\end{IEEEkeywords}

% \fi
\IEEEpeerreviewmaketitle

\section{Introduction}
\label{sec:introduction}

Over the last years, the fast development of data acquisition technologies and the huge growth of multimedia collections (e.g. image, video, music, and others) have made the use of classification systems indispensable~\cite{PaperSurveyCBIR_Datta2008}.
There is a wide range of different applications, including person re-identification~\cite{Karanam2019Survey}, diagnosis of diseases~\cite{paperAlzheimerCBIR}, facial recognition~\cite{bookFaceRecognitionCBIR}, remote sensing~\cite{paperRemoteSensing}, object identification~\cite{paperObjectRecognitionOntologies}, and various others.
However, despite the  significant recent advances in feature extraction methods, effectively retrieving multimedia data still remains a challenge in various scenarios. 
Such complexity is  mainly associated to the diverse aspects involved in human visual perception, which usually can not be encoded by a single visual feature~\cite{paperCBIRRFProb2006,paperPirasFusion17}.

Due to the huge success of deep learning, especially based on Convolutional Neural Networks (CNNs),  multiple models applied to both image and video content,  have been proposed~\cite{paperCNN_RESNET_2016, paperCNN_DPN2017, paperSurveyCNN2014, paperMobileNetv2}.
Despite the remarkable results mainly supported by CNN models~\cite{deepSurvey2020, paperCNN_Alexnet2012, paperCNN_RESNET_2016, paperCNN_DPN2017, paperSurveyCNN2014}, most methods demand a high amount of data to be trained~\cite{sener2018active, deepSurvey2020}.
The availability of supervised training data is often a challenge due to the need for labeling a lot of information, which is expensive and time-consuming~\cite{hino2020active}.
With the objective of easing up the process of labeling data and not requiring manual intervention for this task, there are researches that propose to assist the process of labeling with automatic stages~\cite{sener2018active}.
However, in spite of these possibilities, the process of obtaining labeled data remains a challenge for multiple tasks~\cite{hino2020active},
since the amount of multimedia data available increases much faster than the amount of labeled data that can be obtained for it~\cite{paperReIDApp3}.

In this scenario, various learning paradigms have been attracting increasing attention in order to deal with the scarcity of labeled data.
Unsupervised~\cite{paperCPRR_PRL2017, paperLHRR, paperRDPAC, paperUTAL} and semi-supervised~\cite{Wang_2020_CVPR, paperMandal2020, paperGCN-ICLR2017} strategies often offer attractive solutions.
While the unsupervised approaches require no labeled data at all, the semi-supervised ones require a small set of labeled data.
A recent trend is given by weakly supervised learning~\cite{paperWSIntro}, which is a broad taxonomy that covers different strategies often divided into three main categories: \emph{(i)} incomplete supervision: where only a subset of the training data is labeled, a part of the other subset can be labeled considering active learning or semi-supervised learning; \emph{(ii)} inexact supervision: the available labels are used to create rules and constraints (heuristics) on the training data; and \emph{(iii)} inaccurate supervision: there are wrong or low-quality labels and the idea is to identify the potential mislabeled instances and to correct or remove them.

In general, incomplete supervision often depicts scenarios very close to real-world applications, modeled by Semi-Supervised Learning (SSL) which considers  reliable but limited labeled data. 
Over the last years, SSL approaches have also witnessed huge advances, mainly supported by Graph Convolutional Networks (GCNs)~\cite{paperGCN-ICLR2017}.
Different from traditional CNN models, which generally operate through  convolutions in the Euclidean space, the GCN models allow convolution operations in non-Euclidean domains defined by graph-based structures~\cite{paperGCN-ICLR2017, paperSurveyGCN_2021}.
Although very effective, CNN models often ignore contextual information such as neighborhood references and the relationship between the elements in the dataset.
Furthermore, CNNs are often applied to 2D and 3D data (e.g., images and videos) and are generally not easily applicable to 1D feature vectors, unless some data processing or conversion is done in the original data~\cite{paperFeatureToImageCNN}.

The GCNs exploit multidimensional feature vectors and graph-based neighborhood structures to learn more effective representations.
Due to these aspects, the GCNs have been recently applied for graph-based data on semi-supervised learning tasks, achieving state-of-the-art results.
Several GCN variations have been proposed with relevant results~\cite{paperGCN-GAT-ICLR2018,paperGCN-SGC2019,paperGCN-APPNP2019,paperGCN-ARMA2021}.
The use of GCN has many different applications. There are some recent works that exploit graph learning for question and answer systems~\cite{paperNie2020}, including conversational image search~\cite{paperNie2021}.

While it offers an effective contextual representation learning strategy, GCN models require graph-structured data. The graph data is inherently available in some domains, but needs to be inferred or constructed in others~\cite{paperLDS_GNN_2019}. Consequently, several methods have been proposed  for graph-structured data as citation datasets~\cite{paperGCN-GAT-ICLR2018,paperGCN-SGC2019,paperGCN-APPNP2019,paperGCN-ARMA2021,PaperManLearnGCN_2020,PaperSelfTrain_Li2018,PaperHyperGConv_2019}, but only a few approaches have been proposed for image and multimedia data~\cite{PaperRSIGCN_CVIU2019,PaperGCNReID_2020,PaperGuidedSimRet_NEURIPS2019,PaperGCNRetinalClassif_2021}.
In most cases, the most direct approach is to create a $k$-nearest neighbor graph.
However, the GCN models are sensitive to the input graph, in the sense that a more effective classification depends on the edges between nodes of the same class.

In this paper, we propose a novel GCN-based approach, the Manifold-GCN, for image classification in semi-supervised scenarios, with limited labeled data. 
Deep features are extracted for image representation employing transfer learning by CNNs and Vision Transformers (ViT) models.
Ranking structures are computed and used as input by unsupervised manifold learning algorithms based on these extracted features.
In general, manifold Learning approaches aim to capture and exploit the intrinsic manifold structure to compute a more effective distance/similarity measure~\cite{paperManifoldDef2011}. In this work, we consider recent unsupervised manifold learning methods to provide more effective similarity measures using rank-based formulations.

The manifold learning methods produce more effective ranking results, i.e., improved neighbor sets, which are exploited for building the input graph of the GCN model.
In addition to constructing kNN graphs, the use of reciprocal kNN graphs is proposed.
The main hypothesis of the paper is that the use of manifold learning to improve the graph structure provided as the input of the Graph Convolutional Network (GCN) can further improve the classification results obtained.
This work proposes and validates this hypothesis on different manifold learning and recent GCN approaches.

We can highlight the main contributions of our work as follows: \emph{(i)} novel ways to learn the graph structures that improve GCN image classification; 
\emph{(ii)} the use of reciprocal kNN graph in order to provide a more reliable graph for GCNs.
There are very few works that employ kNN graphs~\cite{paperLDS_GNN_2019} or manifold learning~\cite{PaperManLearnGCN_2020} for GCNs.
In~\cite{paperLDS_GNN_2019} the traditional kNN graph is employed and~\cite{PaperManLearnGCN_2020} uses manifold learning, but in both works no image data is considered.
Other few works have recently employed GCN models on image classification~\cite{PaperRSIGCN_CVIU2019,PaperGCNReID_2020,PaperGuidedSimRet_NEURIPS2019,PaperGCNRetinalClassif_2021}.
However, to the best of our knowledge, this is the first work that exploits both manifold learning and reciprocal kNN graphs for GCN-based semi-supervised image classification.
In addition, it combines powerful contextual modeling given by GCN models with effective representations  given by CNNs and ViT features.

There are many applications of the proposed approach. The improvement of classification results using GCNs may benefit many different areas, especially when there is limited labeled data. For example: person re-identification~\cite{Karanam2019Survey} and diagnosis of diseases~\cite{paperAlzheimerCBIR}. 
The Manifold-GCN can be employed in scenarios where the graph data is not previously available by building the graph from the features and employing manifold learning.

A wide experimental evaluation was conducted in order to assess the effectiveness of the proposed approach.
The experimental results were obtained on 3 public datasets.
We evaluated the impact of different GCN models combined with different manifold learning methods.
The experimental results demonstrate the effectiveness of the proposed approach and the gains on combining manifold learning and reciprocal kNN graphs.

This paper is organized as follows.
Section~\ref{sec:related_work} presents the related work,
while Section~\ref{sec:formaldef} presents the formal definition of semi-supervised learning. Section~\ref{sec:prop_method} describes our proposed approach, the ManifoldGCN.
Section~\ref{sec:gcn_and_ml} presents the GCNs and manifold learning methods considered.
Section~\ref{sec:exp_eval} reports the experimental evaluation.
Finally, Section~\ref{sec:conclusion} states conclusions and considers possible future works.

%............................................................................
\section{Related Work}
\label{sec:related_work}
%............................................................................

This section presents an overview of the methods proposed for semi-supervised image classification over recent years and their main ideas, especially regarding deep learning.

Semi-supervised approaches perform training considering both labeled and unlabeled data, which is advantageous in multiple scenarios where there is little labeled data~\cite{paperSurveySemiSup2019}.
Some of them rely on the generation of pseudo-labels~\cite{paperSurveySemiSupImage2021}. Among the traditional methods for generating pseudo-labels, we can cite: Label Spreading~\cite{Zhou04learningwith} and Pseudo-label~\cite{lee2013pseudo}.
There are also several supervised approaches that later presented semi-supervised variants that do not require the generation of pseudo-labels. For example: Support Vector Machines~\cite{paperSVM} (SVM) and Optimum Path Forest~\cite{paperSSOPF2014} (OPF).

The taxonomy and categories of semi-supervised approaches vary in the literature~\cite{paperSurveySemiSup2019, paperSurveySemiSupImage2021}.
Generally, there is some overlap among categories.
In the following subsections, we present them according to 4 research directions~\cite{paperSurveySemiSupImage2021}: category regularization; stronger augmentation; convergence with self-supervised learning; and graph-based approaches.

\subsection{Consistency regularization}

These methods rely on a concept known as category regularization.
The central idea is to force the approach to produce similar results for augmented versions of the same unlabeled image.
This is generally done by considering an additional term in the loss function.
The first method as far as it is known, to use this concept is called II-Model~\cite{paperIIModel}. In II-Model, they use translation and random horizontal flips as augmentations for unlabeled data, which is often called weak augmentation.

However, the main issue with II-Model is the unstable target, which compromises the algorithm learning procedure.
The Mean Teacher~\cite{paperMeanTeacher} approach was proposed with the intent to address this issue. For this, they use two separate models: the Student network and the Teacher network.
While the Student is trained as usual, the Teacher does not use back-propagation and the weights are updated at each iteration using the weights from the Student network.

\subsection{Stronger Augmentation}

Data augmentation is of crucial importance for various semi-supervised approaches~\cite{paperSurveySemiSupImage2021}. Some strategies focus on improving the performance of classification by employing different kinds of data augmentation techniques, in such a way that the inputs given to the two branches of the neural model (or, to the two separate networks) are sufficiently distinct.
There are many methods that fit in this category, among them: Virtual Adversarial Training and Entropy Minimisation~\cite{paperVAT} (VAT), Unsupervised Data Augmentation~\cite{paperUDA} (UDA), MixMatch~\cite{paperMixMatch}, FixMatch~\cite{paperFixMatch}, ReMixMatch~\cite{paperRemixMatch}, AlphaMatch~\cite{paperAlphaMatch}.
Some of them also mix other ideas, such as the concept of consistency regularization.

\subsection{Convergence with Self-supervised Learning}

Recently, self-supervision has been used by several semi-supervised methods.
Self-supervised approaches are a category of representation learning algorithms capable of generating supervision signals without any human annotations.
Most approaches in this category use self-supervision to generate a set of pseudo-labels for training.
Among the main approaches in this category, we can cite: SimCLR~\cite{paperSIMCLR}, CoMatch~\cite{paperCoMatch}, Self-Match~\cite{paperSelfMatch}.

\subsection{Graph-based Approaches}

A promising research direction is methods based on graphs.
There are different traditional graph-based approaches, both transductive, and inductive ones~\cite{paperSurveySemiSup2019}.
The idea is that the elements of the dataset can be represented as nodes and the edges can be used to propagate or represent some kind of information between these nodes.
Graph-based methods are usually based on the manifold assumption~\cite{paperSurveySemiSup2019}:  the graphs, constructed based on the local similarity between data points, provide a lower-dimensional representation of the potentially high-dimensional input data.
This makes these approaches advantageous for scenarios with data of high dimensionality.

Recently, Graph Convolutional Networks (GCN), have been proposed for semi-supervision.
While CNNs are specially built to operate on regular (Euclidean) structured data, the GNNs work on graphs with different numbers of vertexes and unordered nodes (irregular on non-Euclidean structured data). There are many variants of GCNs proposed: GCN-Net~\cite{paperGCN-ICLR2017}, GCN-SGC~\cite{paperGCN-SGC2019}, GCN-GAT~\cite{paperGCN-GAT-ICLR2018}, GCN-APPNP~\cite{paperGCN-APPNP2019}, GCN-ARMA~\cite{paperGCN-ARMA2021}. Also, variants of GNNs: GNN-LDS~\cite{paperLDS_GNN_2019}, GNN-KNN-LDS~\cite{paperLDS_GNN_2019}.

The GCNs exploit feature vectors and graph-based neighborhood structures to learn more effective representations.
Due to these aspects, the GCNs have been recently applied to graph-based data on semi-supervised learning tasks, achieving state-of-the-art results.
Several GCN variations have been proposed with relevant results~\cite{paperGCN-GAT-ICLR2018,paperGCN-SGC2019,paperGCN-APPNP2019,paperGCN-ARMA2021}.
The use of GCN has many different applications. There are some recent works that exploit graph learning for question and answer systems~\cite{paperNie2020}, including conversational image search~\cite{paperNie2021}.

However, there are still not many approaches for using GCNs in image classification. Among the multiple research topics, there is finding the best approach to model the graph and the features, which are provided as the input for these networks and directly impact their performance and results.

%............................................................................
\section{Graph-based Semi-Supervised Learning Formulation}
\label{sec:formaldef}
%............................................................................

\begin{figure*}[ht!]
    \centering
    \includegraphics[width=.9\textwidth]{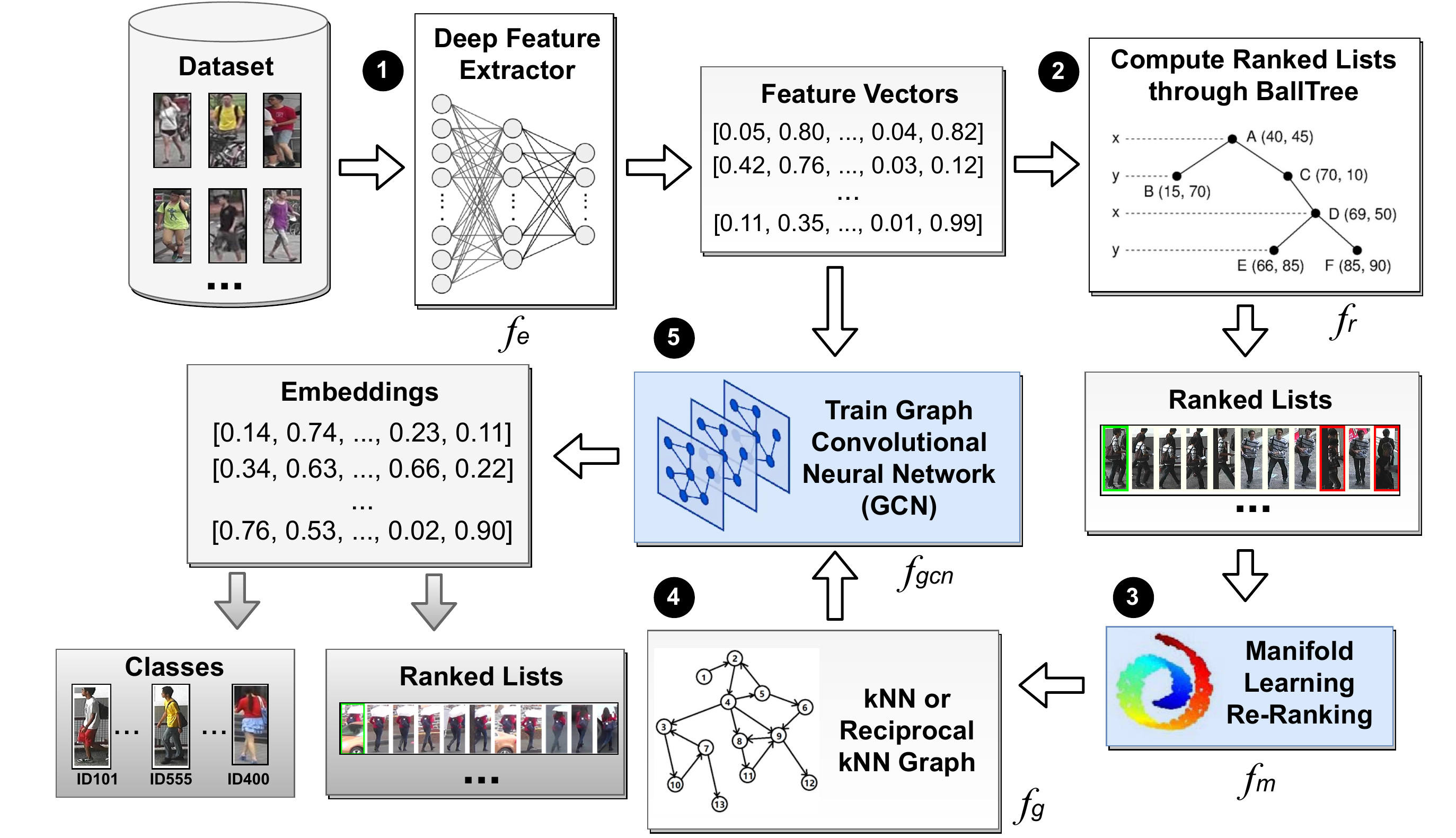}
    \caption{Workflow of our proposed Manifold-GCN framework for image classification. The steps of the approach are numbered.}
    \label{fig:manifold-gcn_steps}
\end{figure*}

In this section, we first discuss a formal definition of the semi-supervised learning setting for classification tasks using GCNs, mostly following the notation from~\cite{paperGCN-ICLR2017,paperDCGP_Latecki_2021}.

Let $\mathcal{C}$=$\{o_{1},$ $o_{2},$ $\dots,o_{n}\}$ be an object collection, where $o_i \in \mathcal{C}$ denotes an image and $n$ denotes the collection size.
The collection is represented by an undirected graph $\mathcal{G}$.
The graph can be formally defined as tuple  $\mathcal{G} = (\mathcal{V}, \mathbf{X}, \mathcal{E})$, where $\mathcal{V}$ denotes the node set,  $\mathbf{X}$ is a feature matrix, and $\mathcal{E}$ denotes the edge set.

The node set is defined by $\mathcal{V} = \{v_1,v_2,\dots,v_n\}$ where each node $v_i \in \mathcal{V}$ represents an image $o_i \in \mathcal{C}$.
Labels can be assigned to nodes $v_i \in \mathcal{V}$, such that a set of labels can be defined as $\mathcal{Y} = \{y_1, y_2, \dots, y_c\}$.
According to the labels, the node set can be more specifically defined as $\mathcal{V} = \{v_1,v_2,\dots,v_L,v_{L+1},\dots,v_n\}$, which denotes a partially labeled data set, where $\mathcal{V}_{L} = \{v_i\}_{i=1}^{L}$ is the labeled data items subset and $\mathcal{V}_{U} = \{v_i\}_{i=L+1}^{n}$ is the unlabeled data items subset.
Formally, the training set can be seen as a labeling function $f_l: \mathcal{V}_L \rightarrow \mathcal{Y}$, where $y_i = f_l(v_i) \forall v_i \in \mathcal{V}_L$. 
In general, on semi-supervised scenarios,  we have $|\mathcal{V}_{L}| \ll |\mathcal{V}_{U}|$.

The feature matrix can be defined as $\mathbf{X} = [\mathbf{x}_1, \mathbf{x}_2, \dots, \mathbf{x}_n]^T \in \mathbb{R}^{n \times d}$, where $\mathbf{x}_i$ is a $d$-dimensional feature vector which represents the image $o_i$, or equivalently, the node $v_i$. 
The vector $\mathbf{x}_i$ is obtained by a feature extraction approach, which can be defined as function $f_e:$ $\mathcal{C} \rightarrow$ $\mathbb{R}^{d}$, such that $\mathbf{x}_i = f_e(o_i)$.

The edge set $\mathcal{E}$ is a set of nodes pairs $(v_i,v_j)$, formally defined as $\mathcal{E} \subseteq \{ (v_i,v_j) | (v_i,v_j) \in \mathcal{V}^2 \wedge v_i \neq v_j \}$.
For graph-structured content, the set $\mathcal{E}$ is intrinsically defined by the data. 
For general image data, we propose to define the set $\mathcal{E}$ based on the feature matrix $\mathbf{X}$. 
How to define an effective graph is a central challenge addressed by our approach, discussed in the next section.

Once defined the graph $\mathcal{G}$, a GCN model denoted by a function $f_{gcn}$ can be used to learn an embedded representation $\mathbf{z}_i$ for each node $v_i$.
The learned representation is exploited to perform classification tasks.
Formally, the classification goal is to learn a function $\hat{f}_l: \mathcal{V}_U \rightarrow \mathcal{Y}$ to predict the labels of unlabeled nodes in $\mathcal{V}_{U}$.

%============================================================
\section{Manifold-based Graph Convolutional Network}
\label{sec:prop_method}
%============================================================

In this work, we propose the Manifold-based Graph Convolutional Network (Manifold-GCN), a semi-supervised framework based on the use of manifold learning and GCN models for image classification for scenarios with limited labeled data.
The initial representations were obtained by deep features extracted by CNN and ViT models trained on a transfer learning setting.
Given the representations, the central idea consists in exploiting contextual similarity measures given by unsupervised manifold learning methods for computing a graph. 
The similarity information encoded in the graph is exploited by GCN models for learning novel representations used for classification.

Figure~\ref{fig:manifold-gcn_steps} illustrates the main steps that compose our strategy. Each step is identified by a number (top of boxes) and a function (bottom of boxes).
In \textbf{(1)}, a feature vector is extracted for representing each image.
In \textbf{(2)}, representations are processed in order to obtain ranked lists, which encode the similarity information.
Unsupervised manifold learning methods are used to analyze contextual similarity information and compute more effective rankings in \textbf{(3)}.
In \textbf{(4)}, the outputs of the manifold learning methods are modeled as kNN graphs or reciprocal kNN graphs.
In \textbf{(5)}, the graph and features are jointly provided to the GCN models for semi-supervised training. 
The embeddings obtained for each of the elements of the dataset can be used for classification, through a softmax operation.
Each of the main steps of the framework is described in the next subsections.

%--------------------------------
\subsection{Similarity Measurement and Ranking Model}
%-------------------------------

In the proposed approach, the similarity information is encoded on ranking structures.
Let us consider a ranking task in which, given a query image, an ordered list of images from the collection is returned according to the similarity to the query. 
Formally, given a query image $o_q$, a ranked list $\tau_{q}$=$(o_{1}$, $o_{2}$, $\dots$, $o_{L})$ in response to the query, where $L$ denotes the length of the list.
The ranked list $\tau_{q}$ can be defined as a permutation of a set $\mathcal{C}_L$ which contains the $L$ most similar images to image $x_q$ in the collection $\mathcal{C}$.
The permutation $\tau_q$ is a bijection from the set $\mathcal{C}_L$ onto the set $[L]=\{1,2,\dots,L\}$.
The $\tau_q (o_i)$ notation denotes the position (or rank) of image $o_i$ in the ranked list $\tau_q$.

The ranked list $\tau_q$ can be computed based on the comparison between image representations.
Let $d$: $\mathbb{R}^{d} \times \mathbb{R}^{d} \rightarrow \mathbb{R}$ be a distance function that computes the distance between two images according to their corresponding feature vectors. The Euclidean distance is often used as the distance function.
Formally, the distance between two images $o_{i}, o_{j}$ is defined by $d$($\mathbf{x}_{i}$, $\mathbf{x}_{j}$).

For a given query, a ranked list can be obtained by sorting images in increasing order of the distance. In terms of ranking positions, we can say that if image $o_{i}$ is ranked before image $o_{j}$ in the ranked list of image $o_q$, that is, $\tau_q(o_i) < \tau_q(o_j)$, then $d(\mathbf{x}_q, \mathbf{x}_i)$ $\leq$  $d(\mathbf{x}_q, \mathbf{x}_j)$.
Taking every image in the collection as a query image $x_{q}$, a set of ranked lists $\mathcal{T}$ = $\{\tau_{1}, \tau_{2},$ $\dots,$  $\tau_{n}\}$ can be obtained.
In this way, the set $\mathcal{T}$ can be obtained from the feature matrix $\mathbf{X}$ and the ranking task defined by a function $f_r$, such that $\mathcal{T} = f_r(\mathbf{X})$.
Tree-based indexing structures~\cite{PaperBallTree} and hashing approaches~\cite{PaperLSH} can be exploited in order to provide efficient implementations for the function $f_r$. In this work, we consider BallTree~\cite{PaperBallTree,scikit-learn} structures. 

%-------------------------------
\subsection{Unsupervised Manifold Learning}
%-------------------------------

How to accurately define distance or similarity among data elements is a challenging and fundamental step in many machine learning tasks. 
The most common approach is given by pairwise comparisons based on Euclidean-like distance functions.
However, pairwise analyses ignore contextual information and complex similarity arrangements encoded in the structural information of the dataset manifold. 
Aiming at addressing such drawbacks, many contextual similarity approaches take into account the structure of datasets in order to compute more global and effective similarity measures.

Manifold Learning is a wide term that has many different definitions in the literature. In general, manifold Learning approaches aim to capture and exploit the intrinsic manifold structure to compute a more effective distance/similarity measure~\cite{paperManifoldDef2011}.
Recently, unsupervised manifold learning approaches based on ranking information have achieved relevant advances in contextual similarity measurement~\cite{paperLHRR,paperRDPAC,paperBFSTREE}.

In fact, the set of ranked lists $\mathcal{T}$ encodes rich similarity information about the image collection.  
The main objective of rank-based manifold learning methods is to exploit such information to capture the structure of the dataset manifold. 
Therefore, this step consists of the use of unsupervised manifold learning methods for processing the original ranked lists, providing more effective ranking results which  are subsequently modeled as graphs to be submitted to a GCN model.

Formally, the manifold learning methods can be defined as a function $f_{m}$ that receives  a set of ranked lists $\mathcal{T}$ as input and returns a set of ranked lists $\mathcal{T}_{m}$ as output, which is expected to be more effective than the original:

\begin{equation}
    \mathcal{T}_m = f_{m}(\mathcal{T}).
\end{equation}

Once defined under a common formulation, three different manifold learning algorithms were considered to instantiate the proposed approach (described in Section~\ref{sec:gcn_and_ml}).

%--------------------------------
\subsection{Graph Building}
%-------------------------------

The improved set of ranked lists computed by the manifold learning methods is used to build a graph. 
The motivation is based on the conjecture that more effective similarity information can be extracted and encoded in the graph by exploiting the processed ranked lists.
Let $\mathcal{G} = (\mathcal{V}, \mathbf{X}, \mathcal{E})$ be the graph defined in Section~\ref{sec:formaldef}.
We propose to compute the edge set $\mathcal{E}$ as a function of the set of ranked lists $\mathcal{T}_m$, such that $\mathcal{E} = f_g(\mathcal{T}_m)$.

This work considers two distinct approaches to define the function $f_g$.
The similarity information encoded in the ranked lists is modeled through different neighborhood set formulations.
Both approaches are discussed in the following.

$\bullet$ \textit{Traditional kNN Graph:} the kNN graph is based on the natural neighborhood set.
Given an element  $o_{q}$, the natural neighborhood set $\mathcal{N}(o_q, k)$  contains the $k$ most similar elements to $o_{q}$, which can be formally defined as:

\begin{equation}
\begin{split}
\mathcal{N}(o_q, k) = \{ \mathcal{X} \subseteq \mathcal{C}, |\mathcal{X}| = k~\wedge~\forall o_{i} \in \mathcal{X}, o_{j} \in \mathcal{C} - \mathcal{X} : \\ \tau_{q}(o_i) < \tau_{q}(o_j) \}.
\end{split}
\end{equation}
Therefore, the edge set $\mathcal{E}$ of the kNN graph can be defined as:

\begin{equation}
\mathcal{E} =  \left\{ (o_{q}, o_{j}) ~|~ o_{j} \in \mathcal{N}(o_q, k) \right\}.
\end{equation}
In other words, each element has an edge to the $k$ most similar elements.

$\bullet$  \textit{Reciprocal kNN Graph: } the reciprocal kNN graph is based on the reciprocal neighborhood set~\cite{paperHelloNeighbor2011}, which requires a stronger bidirectional similarity relationship.
Different from the natural neighborhood set, which is not symmetrical, the reciprocal neighborhood set is symmetrically defined as:
\begin{equation}
\mathcal{N}_{r}(o_q, k) = \left\{ obj_{i} | obj_{i}  \in \mathcal{N}(o_q, k) \wedge o_{q} \in \mathcal{N}(o_i, k) \right\}.
\end{equation}
The edge set $E$ for the reciprocal kNN set can be defined as:
\begin{equation}
\mathcal{E} =  \left\{ (o_{q}, o_{j}) ~|~ o_{j} \in \mathcal{N}_{r}(o_q, k) \right\}.
\end{equation}
Thus, we can interpret that there are edges between the elements $o_{q}$ and $o_{j}$ if they are reciprocal neighbors in the top-$k$ positions of their ranked lists.

For both kNN and reciprocal kNN approaches, the edge set $\mathcal{E}$ can be represented by a non-negative adjacency matrix $\mathbf{A} = [a_{ij}] \in \mathbb{R}^{n \times n} $, which can be defined as:

\begin{equation}
        a_{ij} = 
\begin{cases}
    1,              & (o_i,o_j) \in \mathcal{E} \\
    0,              & \text{otherwise.}
\end{cases}
\end{equation}
The adjacency matrix $\mathbf{A}$ is used as input by GCN models, as discussed in next section.

%............................................................................
\subsection{Graph Convolutional Networks}	
\label{sec:GCN}
%............................................................................

Graph Convolutional Networks (GCN), originally introduced in~\cite{paperGCN-ICLR2017}, aim at learning novel and more effective representations (embeddings) for each graph node.
It is done by iteratively aggregating the embeddings of its neighbors, encoding the graph structure directly in a neural network model.
The original model proposed in~\cite{paperGCN-ICLR2017} is a two-layer GCN model which uses the graph represented by the  adjacency matrix $\mathbf{A}$ for semi-supervised node classification. 

The network model can be depicted as a function both on the feature data $\mathbf{X}$ and on the adjacency matrix $\mathbf{A}$, as:

\begin{equation}
  \mathbf{Z} = f_{gcn} (\mathbf{X}, \mathbf{A}), 
\end{equation}

\noindent where $\mathbf{Z}$ denotes an embedding matrix, such that 
$\mathbf{Z} =$ $[\mathbf{z}_1, \mathbf{z}_2,$ $\dots,$ $\mathbf{z}_n]^T \in \mathbb{R}^{n \times c}$ and $\mathbf{z}_i$ is a $c$-dimensional embedded representation learned for the node $v_i$; where $n$ is the dataset size and $c$ corresponds to the number of classes.

The degree matrices are computed as a pre-processing step, defined as $\hat{\mathbf{A}} = \tilde{\mathbf{D}}^{-1/2} \tilde{\mathbf{A}} \tilde{\mathbf{D}}^{-1/2}$, where $\tilde{\mathbf{A}} = \mathbf{A} + \mathbf{I}$ and $\tilde{\mathbf{D}}$ is the degree matrix of $\tilde{\mathbf{A}}$.
Then, the function  $f_{gcn}(\cdot)$ which represents the two-layer GCN model assumes the form:

\begin{equation}
\mathbf{Z} = f(\mathbf{X}, \mathbf{A}) = softmax ( \hat{\mathbf{A}} ~ ReLU (\hat{\mathbf{A}} \mathbf{X} \mathbf{W}^{(0)} ) \mathbf{W}^{(1)} ).
\end{equation}

The matrix $\mathbf{W}^{(0)} \in \mathbb{R}^{d \times H}$ defines the neural network weights for an input-to-hidden layer with $H$ feature maps, while $\mathbf{W}^{(1)} \in \mathbb{R}^{H \times c}$ is a hidden-to-output matrix.
Both matrices $\mathbf{W}^{(0)}$ and $\mathbf{W}^{(1)}$ are trained using gradient descent, considering the cross-entropy error over all labeled nodes. $v_l \in \mathcal{V}_L$.

The activation function is applied row-wise and is defined as $softmax(z_{i}) = \frac{exp(z_{i})}{\sum_{i} exp(z_{i})}$, where $z_{i}$ is the position $i$ of embedding $\mathbf{z}_{i}$.

The softmax yields the probability distribution over the $c$ class labels for each row, i.e., the probability values sum up to 1 for each row. 
Given an image $o_i$, the learned embedded representation $\mathbf{z}_i$ is then used for classification tasks by applying an argmax over the output of the softmax.

%-------------------------------
\section{GCNs and Manifold Learning Methods}
\label{sec:gcn_and_ml}
%-------------------------------

The proposed approach is flexible in the sense that it can be instantiated by different GCN models and manifold learning methods.
This section briefly describes the GCN models and the manifold learning methods considered in this work.

%-------------------------------
\subsection{GCN Models}
%-------------------------------

The original GCN~\cite{paperGCN-ICLR2017} model and more 4 variants~\cite{paperGCN-APPNP2019,paperGCN-SGC2019,paperGCN-ARMA2021} are used in the proposed Manifold-GCN approach.
The GCN models employed are:

\begin{itemize}

\item \textbf{Graph Convolution Network (GCN)}~\cite{paperGCN-ICLR2017}: the first GCN proposed, introducing the idea of convolutions applied to graph domains, often known as GCN-Net or simply GCN;

\item \textbf{Simple Graph Convolution (SGC)}~\cite{paperGCN-SGC2019}: a simplification of the conventional GCN models which removes the non-linearities and collapses weight matrix between consecutive layers;

\item \textbf{Graph Attention Networks (GAT)}~\cite{paperGCN-GAT-ICLR2018}: employs auto-attention layers with the idea of solving the main shortcomings of the previous GCN models. The layers are stacked in a way that it is possible to specify different weights for nodes of the same neighborhood without requiring costly operations;
 
\item \textbf{Approximate Personalized Propagation of Neural Predictions (APPNP)}~\cite{paperGCN-APPNP2019}: a model that combines a GCN with the PageRank algorithm, deriving a propagation strategy based on a modified PageRank approach;

\item \textbf{Auto-Regressive Moving Average (ARMA) Filter Convolutions}~\cite{paperGCN-ARMA2021}: a GCN variant that defines convolutional layers based on filters of Auto-Regressive Moving Average type.

\end{itemize}

%-------------------------------
\subsection{Manifold Learning Methods}
%-------------------------------

Manifold learning can be broadly understood as the process of non-linear dimensionality reduction by performing distance learning for a set of features.
In fact, images are commonly represented as points in a high-dimensional feature space.
However, it has been shown that data samples often live in a much lower dimensional intrinsic space~\cite{paperManifoldDef2011}.
Therefore, how to capture and exploit the intrinsic manifold structure to compute a more effective distance/similarity measure becomes a key task in many areas~\cite{paperManifoldDef2011}.
In this work, we consider recent unsupervised manifold learning methods to provide more effective similarity measures using rank-based approaches.
Three of them are considered:

\begin{itemize}
%--------------------------------
\item \textbf{Log-based Hypergraph of Ranking References (LHRR)}~\cite{paperLHRR}: an algorithm that models the input ranked lists as hypergraphs and exploits the relations between the elements in the dataset.
%--------------------------------
\item \textbf{BFS-Tree of Ranking References (BFSTREE)}~\cite{paperBFSTREE}: it uses a breadth-first tree structure that models the similarity information between the elements in the ranked lists, which is employed with the objective of analyzing the implicit relations between the elements of the dataset. The tree structure allows a representation of the top-$k$ elements such that the weights of the edges are computed based on the correlations among the ranked lists.

%--------------------------------
\item \textbf{The Rank-based Diffusion Process with Assured Convergence (RDPAC)}~\cite{paperRDPAC}: it performs a diffusion process to exploit the information contained in the ranked lists.
It also presents formal proof for the convergence of the diffusion process.
The asymptotic complexity of the algorithm is low, which allows its use in many different scenarios with a great number of data elements.

\end{itemize}

%*****************************************************
\section{Experimental Evaluation}
\label{sec:exp_eval}
%*****************************************************

This section discusses the experimental evaluation conducted to assess the effectiveness of the proposed Manifold-GCN.
Section~\ref{ssec:datasets} describes the datasets and features considered. Section~\ref{sec:prot_exp} discusses the experimental protocol.
The semi-supervised image classification results are presented in Section~\ref{ssec:classifresults}.
Section~\ref{ssec:visresults} shows  visualizations of feature space improvements,
while Section~\ref{ssec:comparison} reports a comparison with both traditional and recent state-of-the-art methods,
Section~\ref{ssec:effeciencyresults}  reports the run-time for each step of the proposed approach.

%-------------------------------------------------------
\subsection{Datasets and Features}
\label{ssec:datasets}
%-------------------------------------------------------

Three public datasets were considered in the experimental evaluation:

\begin{itemize}
\item \textbf{Flowers17~\cite{PaperFlowers}}: traditional dataset composed of 1,360 images of 17 species of flowers;
\item \textbf{Corel5k~\cite{PaperCorel5k_PR2013}}: 5,000 diverse images (cars, animals, buildings, and others) divided into 100 classes;
\item \textbf{CUB-200-2011~\cite{WahCUB_200_2011}}: a popular benchmark for image classification composed by 11,788 photos of 200 bird species;
\end{itemize}

A diverse set of deep features was considered in the conducted experiments.
For the general purpose datasets (Flowers17, Corel5k, and CUB-200-2011), all CNNs were trained on ImageNet~\cite{paperImagenet2009} dataset through. The extractions were performed with the PyTorch framework~\footnote{\url{https://github.com/Cadene/pretrained-models.pytorch}}.
In all cases, the Euclidean distance was considered.

We also employ two different types of Vision Transformers: An Image is Worth 16x16 Words Visual Transformer (VIT-B16)~\cite{paperVIT16}~\footnote{\url{https://github.com/faustomorales/vit-keras}} and Tokens-To-Token Vision Transformer (T2T-VIT)~\cite{paperT2T}~\footnote{\url{https://github.com/yitu-opensource/T2T-ViT}}.
Both were trained on ImageNet~\cite{paperImagenet2009} dataset.

%-------------------------------------------------------
\subsection{Experimental Protocol}
\label{sec:prot_exp}
%-------------------------------------------------------

In this work, the training data is the labeled set and the testing data is the unlabeled set.
The unlabeled data considered during the training process comes from the test set.
This protocol was also adopted for all baselines.

For the manifold learning approach, all the data is used for the distance learning process, which is completely unsupervised; no labels are used.
In the second step, the semi-supervised classification by the GCN, we perform cross-validation that, in our case, consists of a 10-fold split where one fold is used for training and the rest is used for testing. 
For each of 10 executions (one for every fold being considered as training) and 90\% is considered as testing data (unlabeled data).
We highlight that, since we are running 10 executions by changing the folds, every dataset element will be considered as training or test at least once.
Therefore, each reported value corresponds to the mean of 50 executions (number of executions multiplied by the number of folds).

For all the GCNs, the Adam optimizer with a learning rate of $10^{-5}$ was used, except for Cub200, in which we used a learning rate of $10^{-4}$. Regarding the number of neurons, we used 256. The only exceptions are GCN-SGC, which does not have this parameter; and GCN-GAT which has a number of heads, which was set to 32. The training processes consisted of 200 epochs, using input graphs with $k=40$.
In the same way, the manifold learning methods also have a parameter $k$, which is different from the graph $k$. For the method $k$, we also used $k=40$.

\subsection{Classification Results}
\label{ssec:classifresults}

The proposed approach was evaluated on a wide diversity of semi-supervised classification scenarios, considering 3 distinct datasets (Flowers, Corel5k, and CUB200).
For each dataset,  4 to 5 deep learning features trained on a transfer learning setting were used, considering both CNNs and Vision Transformers approaches.
 For classification, 5 GCNs models are evaluated considering both the traditional and reciprocal kNN graphs. 
 The impact of re-ranking step is also assessed, evaluating the classification results with and without this step, considering 3 distinct rank-based manifold learning methods.
In the semi-supervised scenario, the mean of 5 executions for 10 folds was performed.

Tables~\ref{tab_gcn_flowers},~\ref{tab_gcn_corel5k} and~\ref{tab_gcn_cub200} present the results for the datasets Flowers, Corel5k, and CUB200, respectively.
The best result for each feature/GCN is highlighted in bold.
The gray highlight is used to indicate the best result for the corresponding GCN.
The blue color indicates the best result for the dataset (the best result in the whole table).

Some interesting observations can be made from the experimental results.
In general, it can be noticed that the reciprocal kNN graph outperforms the traditional kNN graph. It can be observed that the use of manifold learning methods outperforms the scenarios without its use.
Moreover, the combination of reciprocal kNN graph and manifold learning methods leads to the best results for all GCN models (gray highlight) and datasets (in blue).

Among the features, VIT-B16 yielded the best results.
Therefore, there is a correlation that shows that the better the feature, the better the classification result.
In this case, the best feature is VIT-B16.
For GCN models and manifold learning methods, the diversity is higher, but GCN-SGC and RDPAC achieved the best results in most of the scenarios.
We also can highlight the remarkable gains obtained on all datasets and features from the original (kNN without re-ranking) to the proposed approach (reciprocal kNN with re-ranking). For CUB200, the most challenging dataset, the accuracy of GCN-APPNP was improved from 55.24\% to 75.59\%.

\begin{figure*}[!htb]
    \centering
    \begin{tabular}{cc} 
    \includegraphics[width=.49\textwidth]{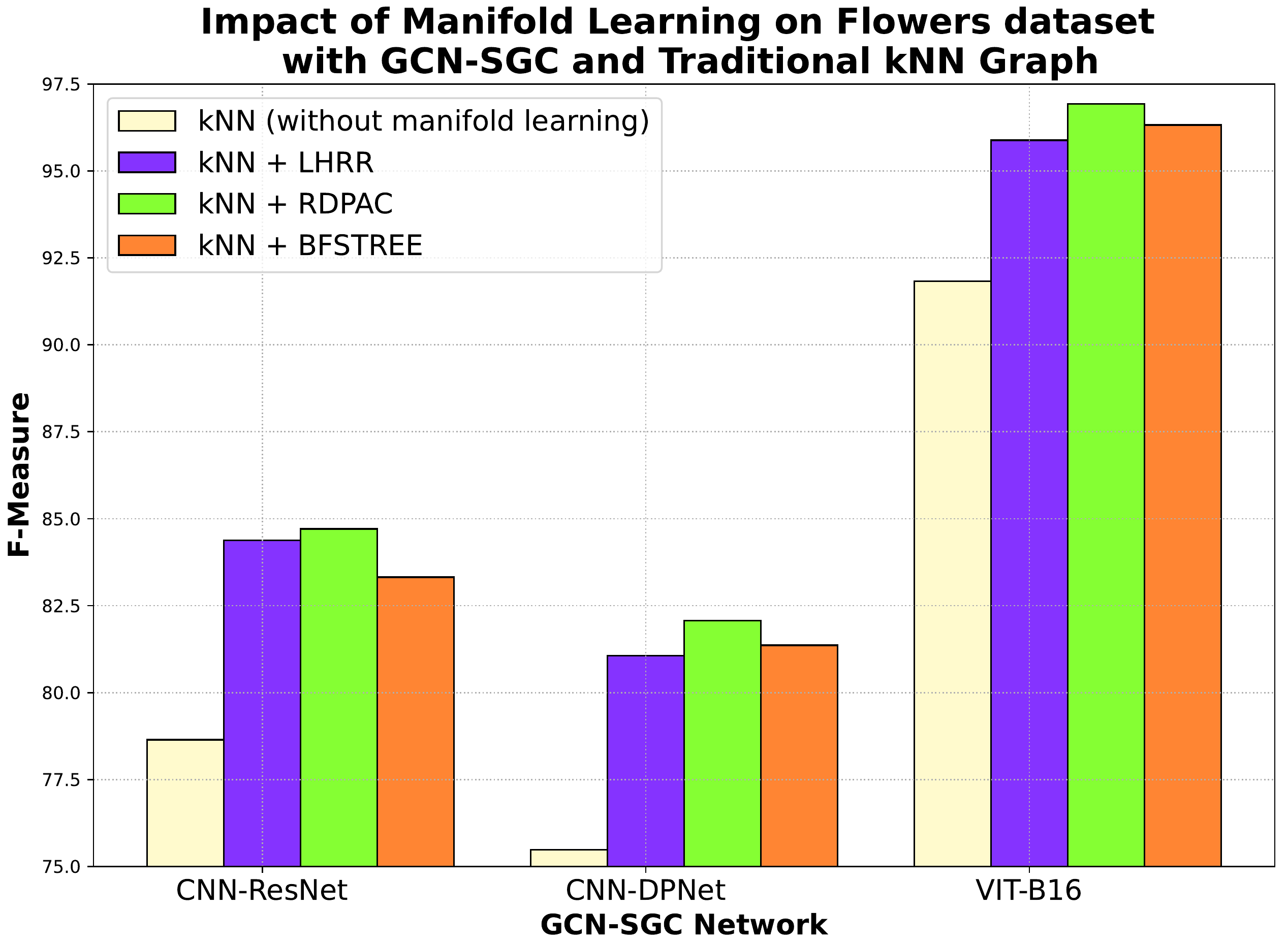} &  \includegraphics[width=.49\textwidth]{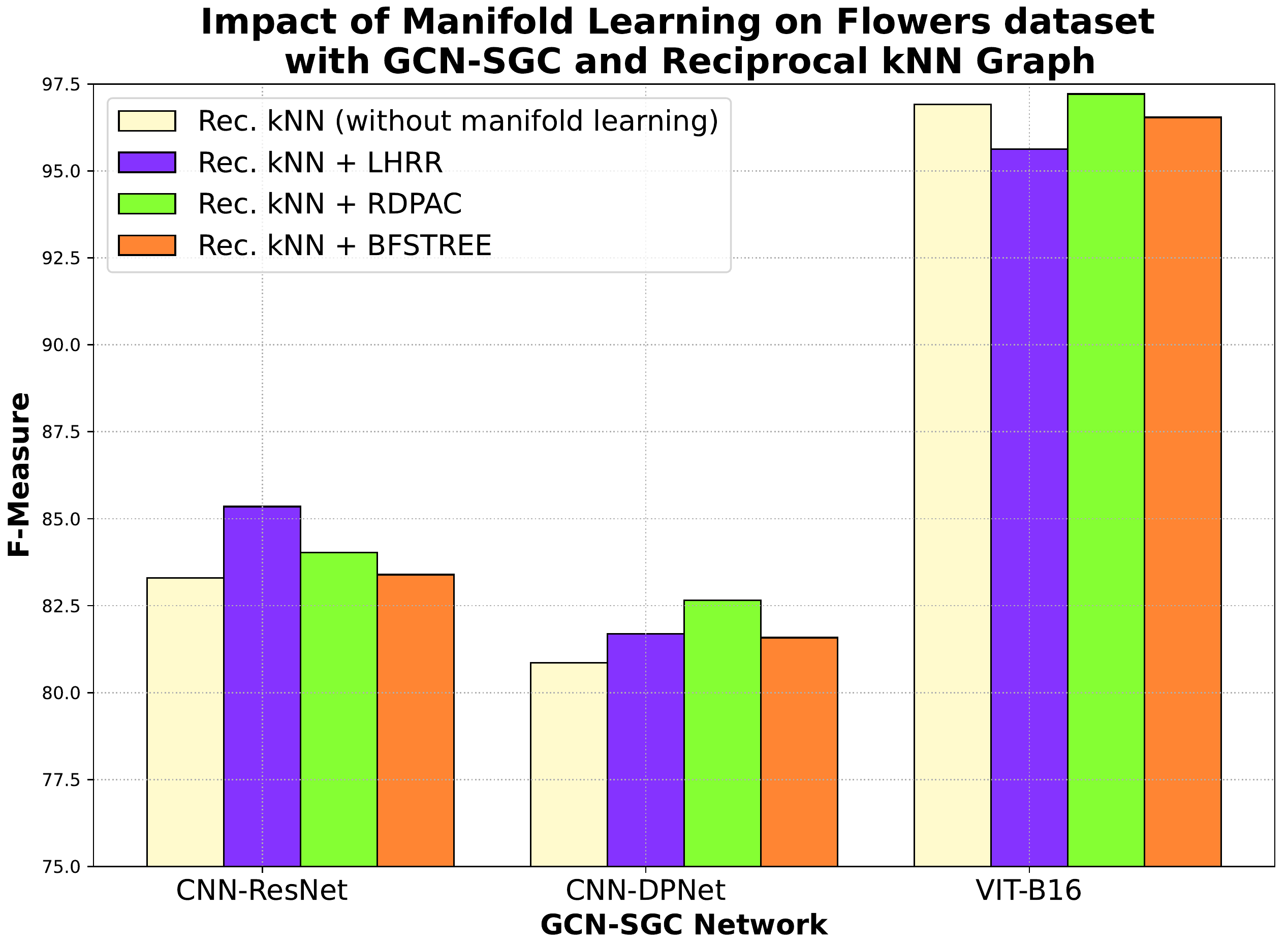} \\
    (a) \textbf{Flowers - Traditional kNN} & (b) \textbf{Flowers - Reciprocal kNN} \\ \\  
    \includegraphics[width=.49\textwidth]{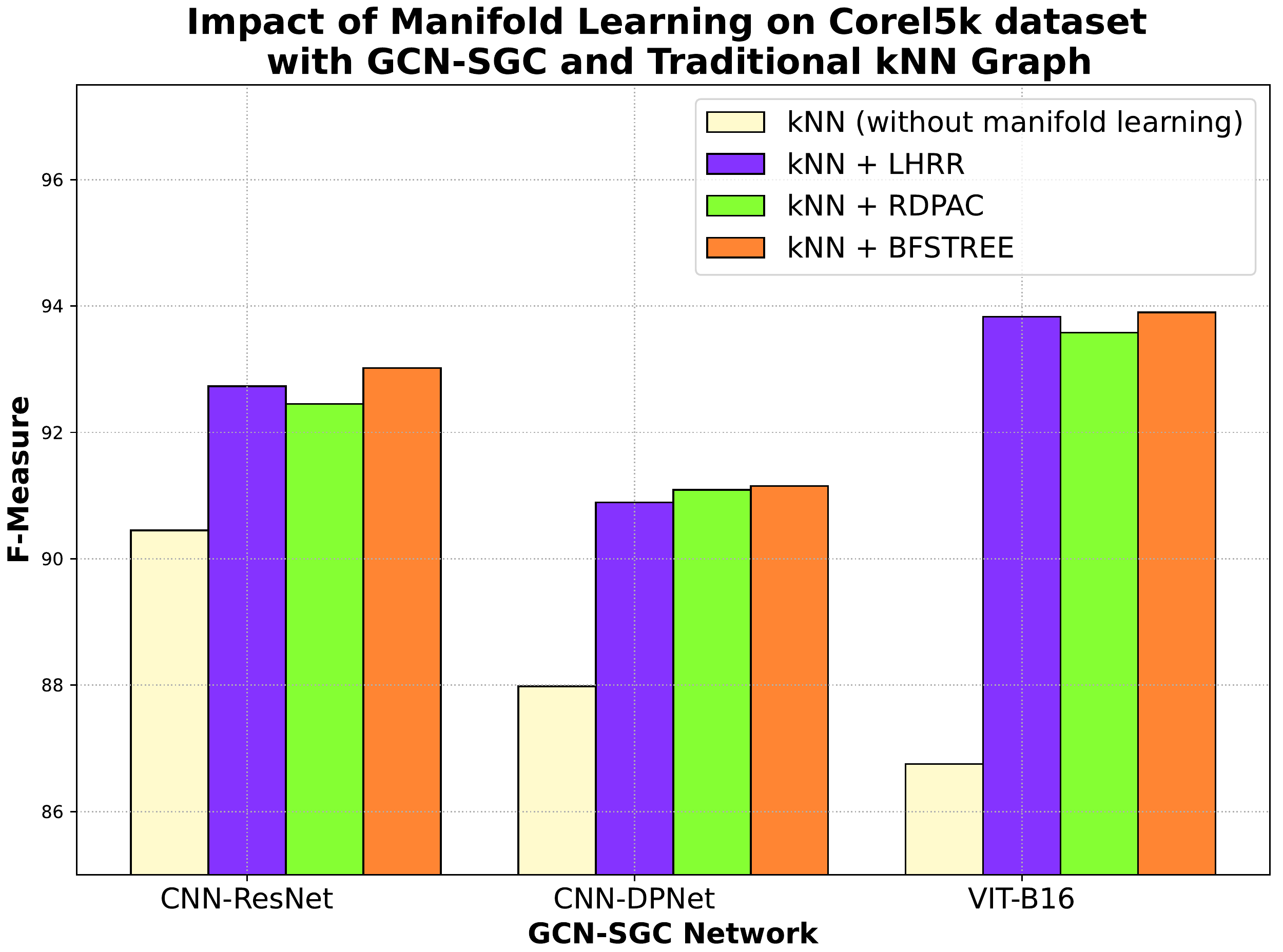} &  \includegraphics[width=.49\textwidth]{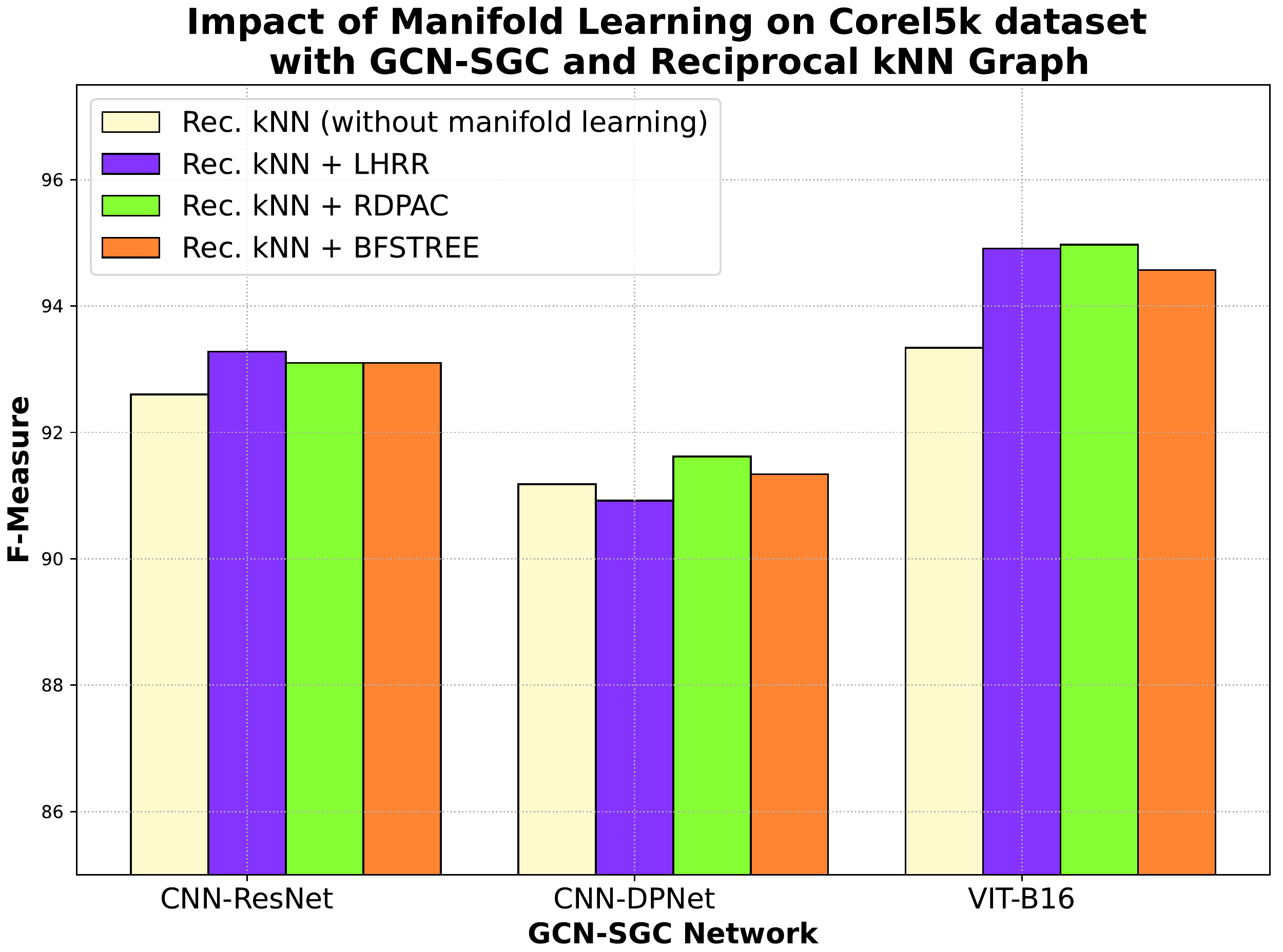} \\
    (c) \textbf{Corel5k - Traditional kNN} & (d) \textbf{Corel5k - Reciprocal kNN} \\ \\  
    \includegraphics[width=.49\textwidth]{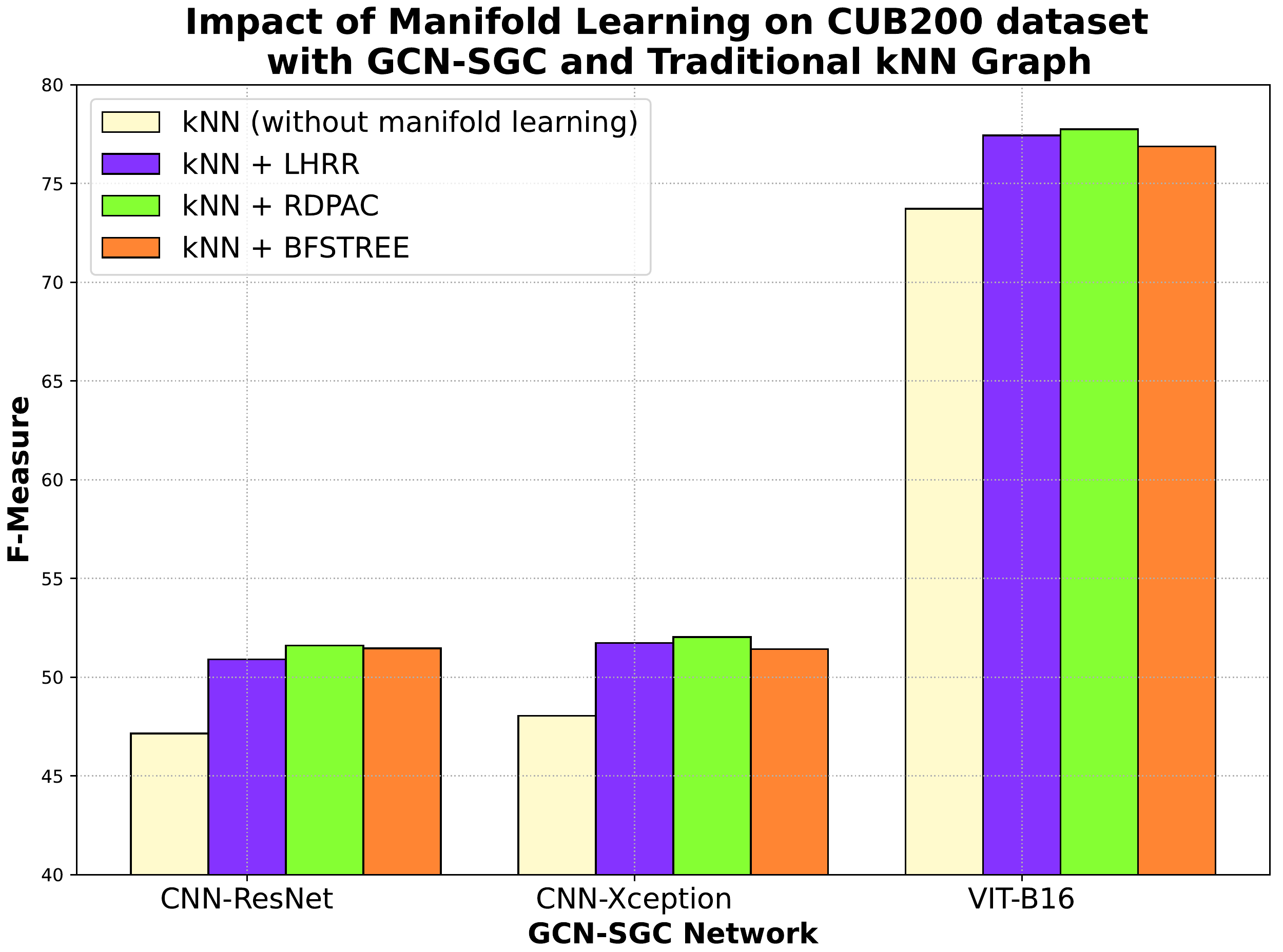} &  \includegraphics[width=.49\textwidth]{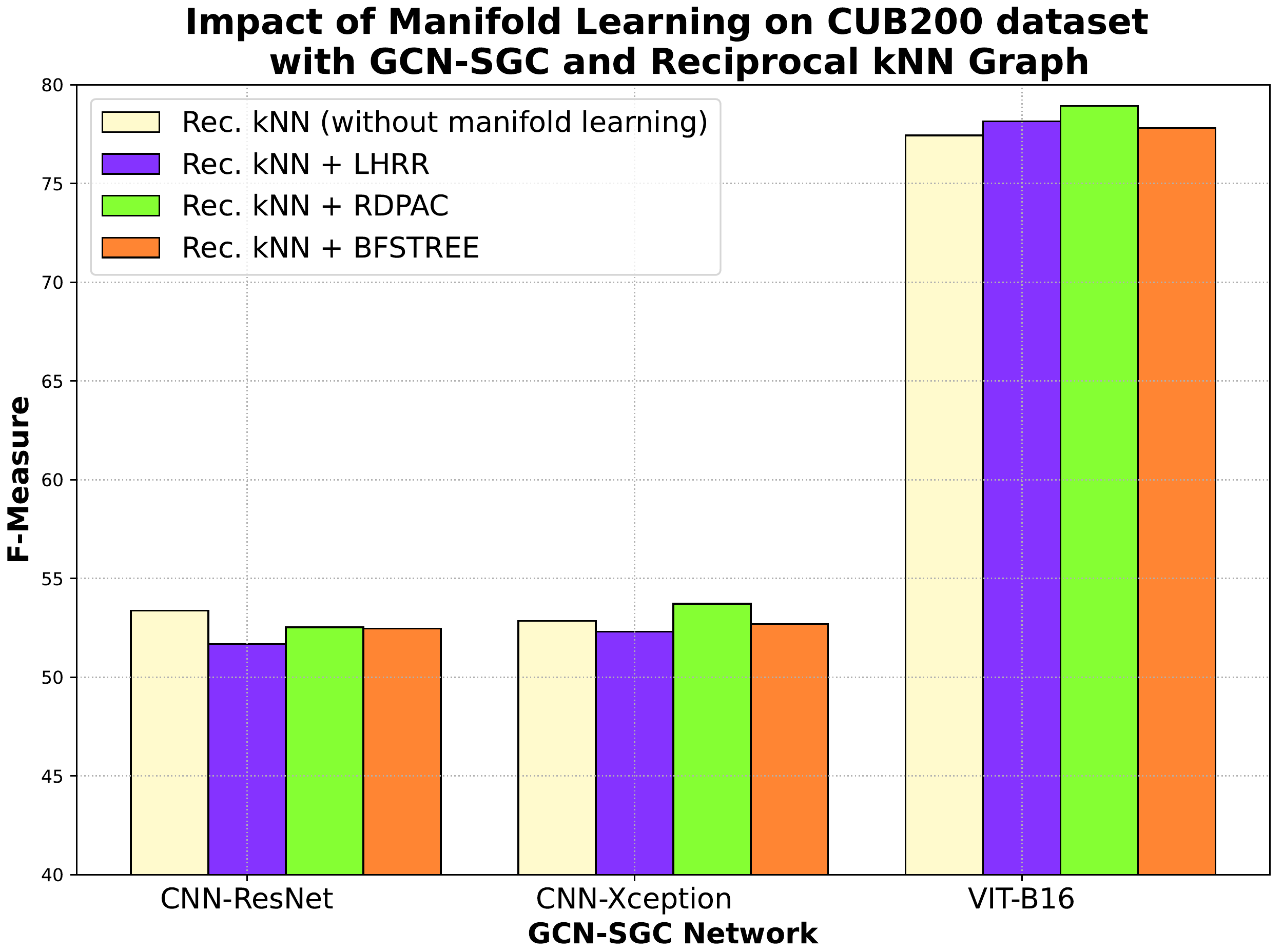} \\
    (e) \textbf{Cub200 - Traditional kNN} & (f) \textbf{Cub200 - Reciprocal kNN} \\ \\ 
    \end{tabular}
    \caption{Impact of manifold learning approaches on F-measure results considering GCN-SGC on different datasets and features.}
    \label{fig:graphs_fmeasure}
\end{figure*}

Our method was also evaluated considering the weighted F-Measure. Figure~\ref{fig:graphs_fmeasure} reports the results for GCN-SGC on the traditional kNN graph (on the left) and the Reciprocal kNN graph (on the right). For every graph, we see that using manifold learning improves the results of the traditional GCN.

\subsection{Visualization Results}
\label{ssec:visresults}

\begin{figure*}[!htb]
    \centering
    \begin{tabular}{cc} 
    \includegraphics[width=.48\textwidth]{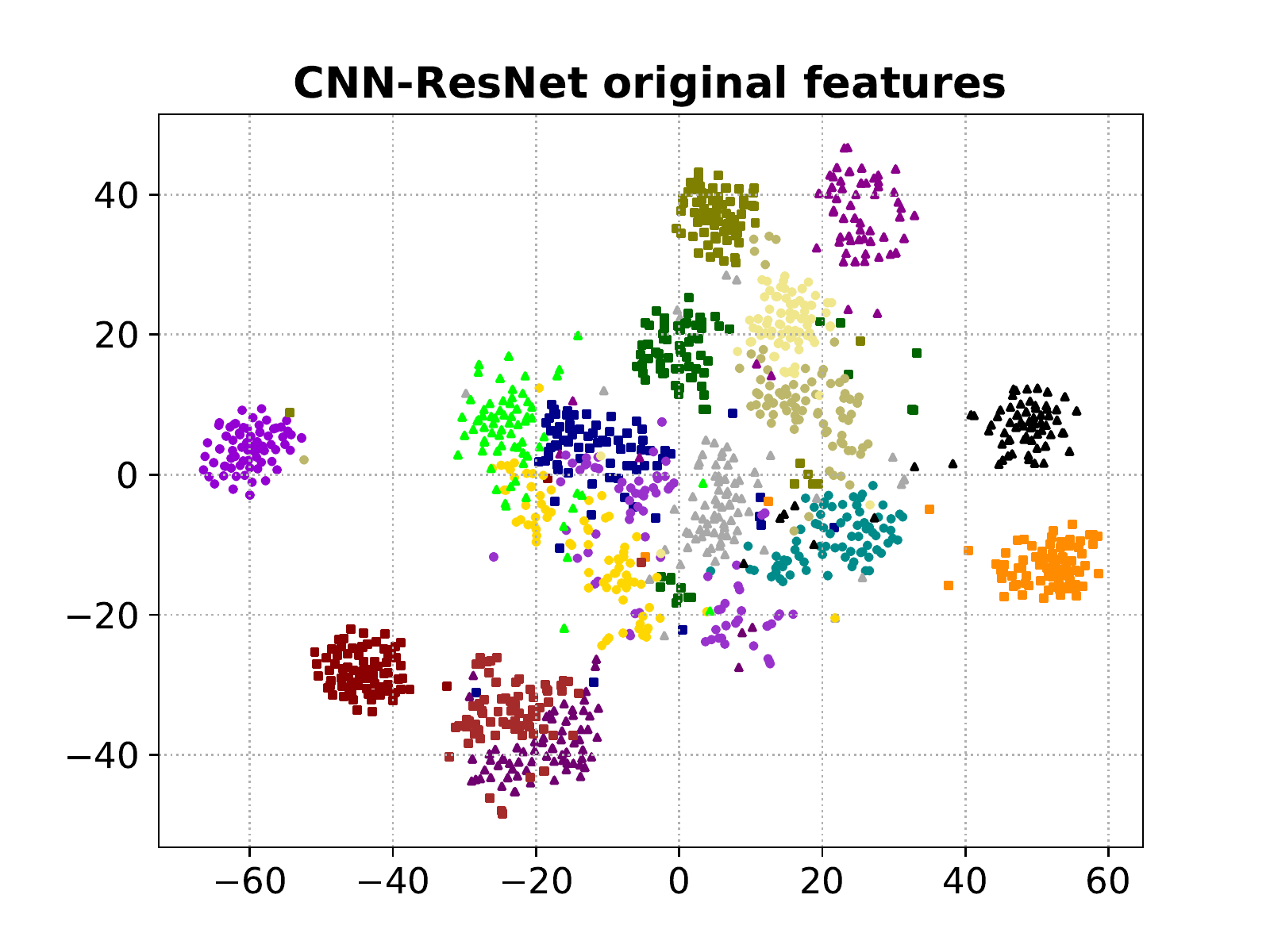} &  \includegraphics[width=.48\textwidth]{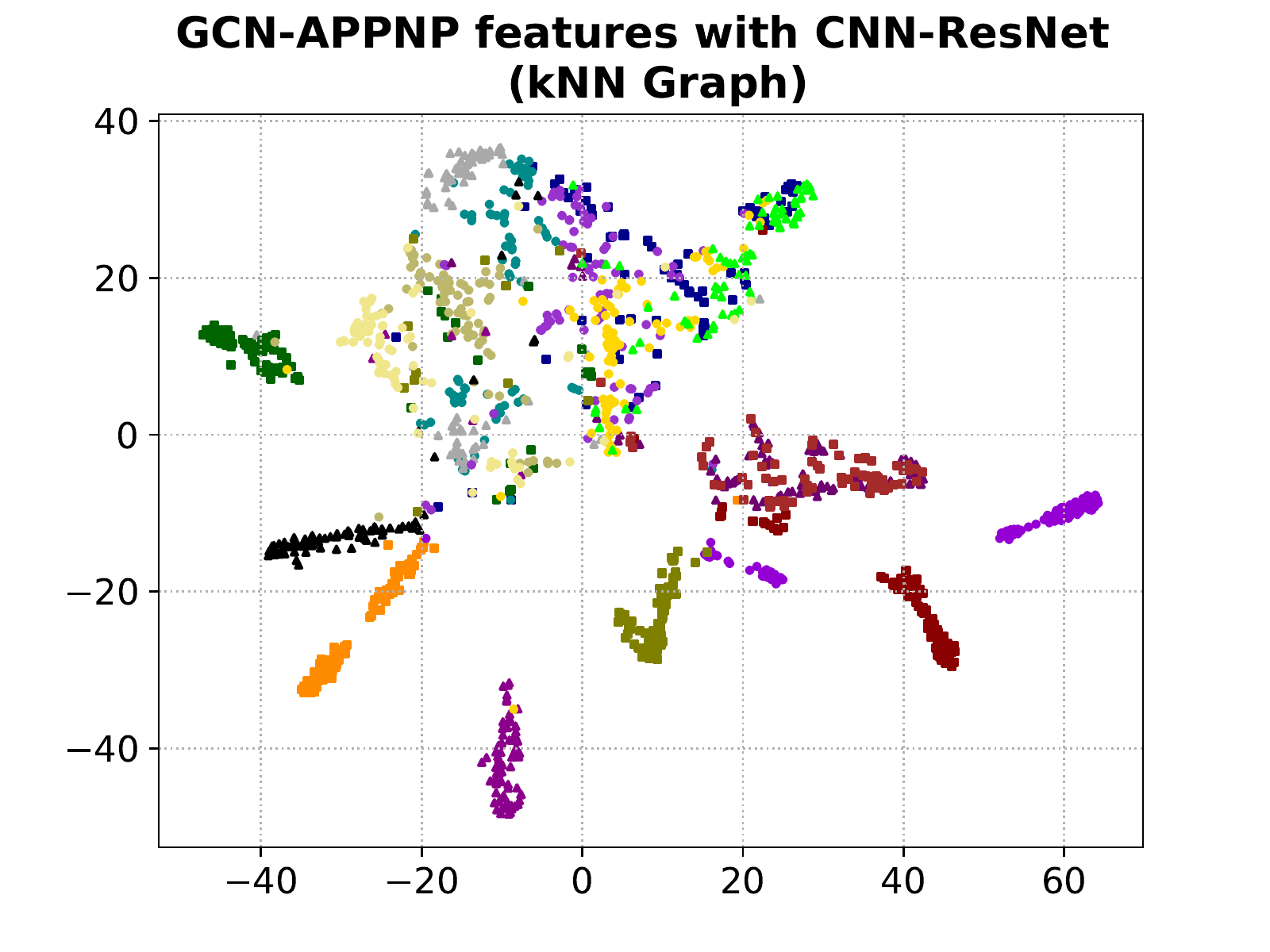} \\
    (a) \textbf{Original Features} & (b) \textbf{GCN-APPNP Features (kNN Graph)} \\ \\  
    \includegraphics[width=.45\textwidth]{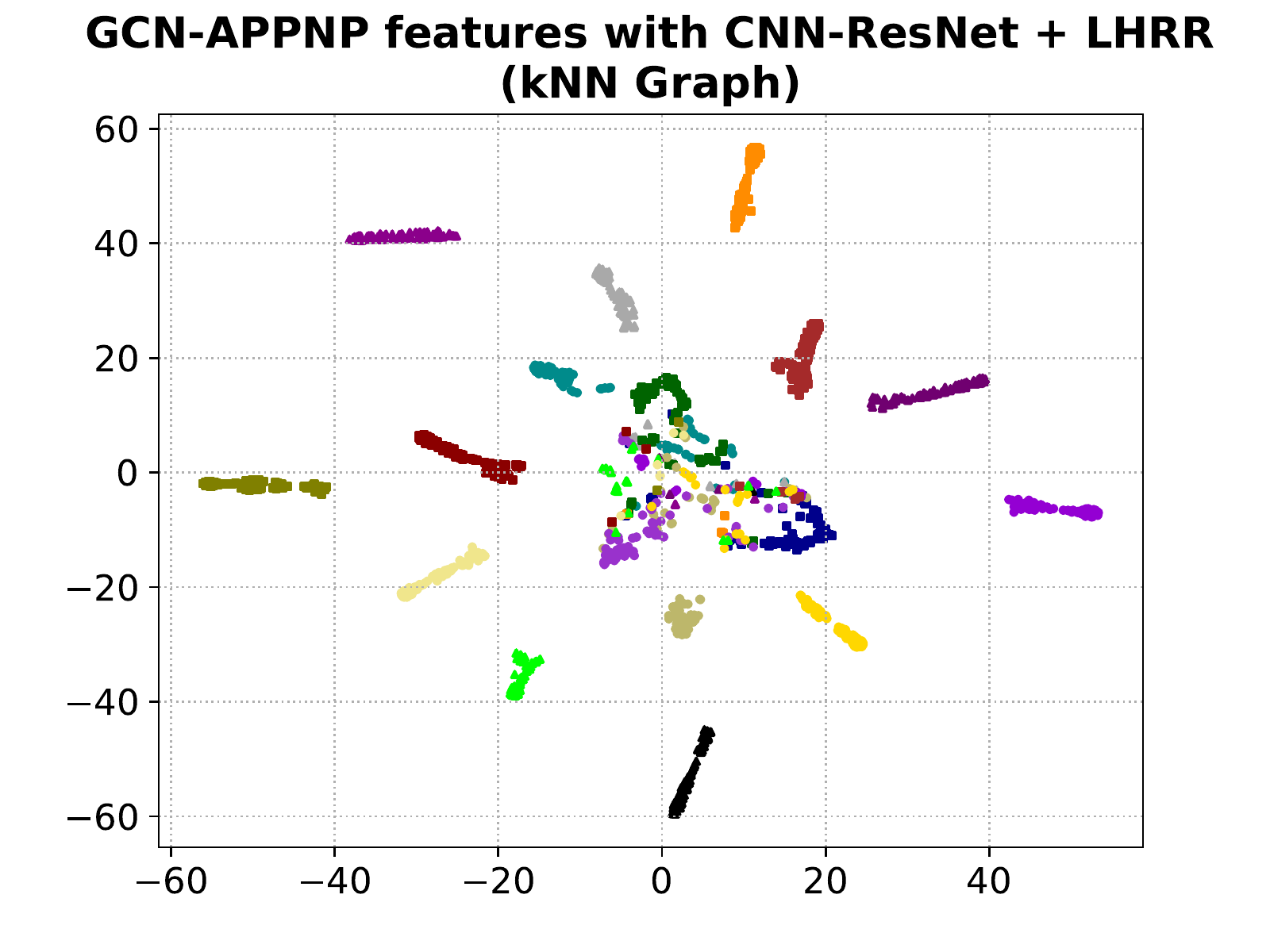} &  \includegraphics[width=.48\textwidth]{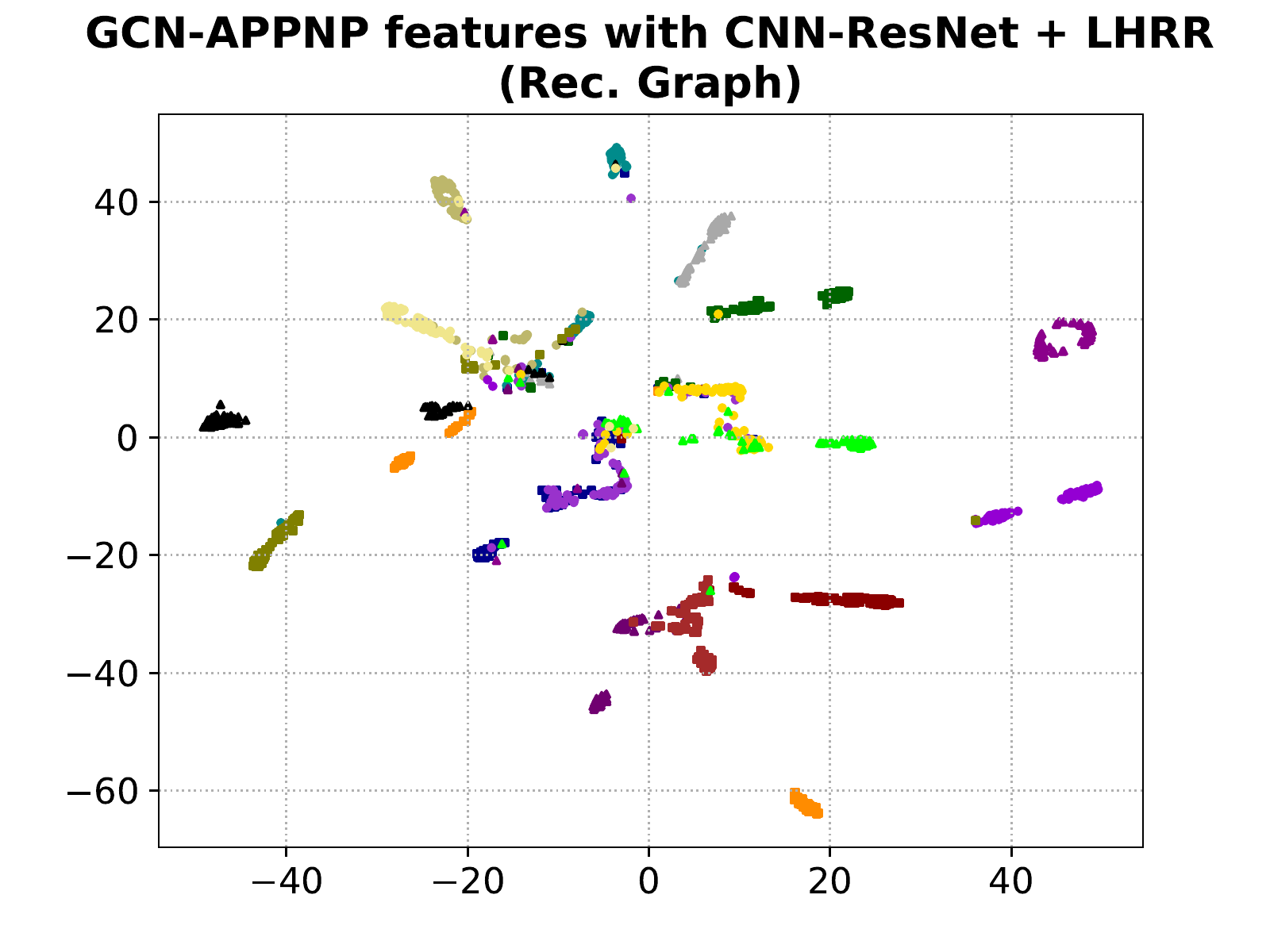} \\
    (c) \textbf{GCN-APPNP Features with LHRR (kNN Graph)} & (d) \textbf{GCN-APPNP Features with LHRR (Rec. Graph)} \\ \\  
    \end{tabular}
    \caption{t-SNE visualizations that show the feature space improvement when manifold learning and reciprocal graph were applied. Experiments were conducted on the Flowers dataset and CNN-ResNet features. Each class is represented by a different shape and color.}
    \label{fig:graphs_tsne}
\end{figure*}

In order to visualize the effectiveness of our approach, an experiment was conducted showing the distribution of features in a 2D space, after being processed by t-Distributed Stochastic Neighbor Embedding (TSNE)~\cite{paperTSNE}.
Figure~\ref{fig:graphs_tsne} shows the results for (a) the original CNN-ResNet features; (b) the GCN output with kNN graph; (c) the GCN output with kNN graph and manifold learning; (d) the GCN output with Reciprocal graph and manifold learning.
The Flowers-17 was chosen for this visualization due to the small number of classes, which makes it easier to visualize the improvements.
Each class is represented by a different combination of shape and color.
Notice that the distribution of classes is further improved when the Manifold-GCN is applied (c and d), which is consistent with our main hypothesis.

\begin{table*}[!t]
\caption{Impact of manifold learning approaches (LHRR, RDPAC, BFSTREE) and Reciprocal Graph (Rec.) on the classification accuracy (\%) of 5 different GCN models on Flowers17 dataset. The best results for each feature and GCN model are highlighted in bold, the best results for each GCN model are marked with a gray background, and the best result for the entire dataset is highlighted in blue. In all the cases, the best results used manifold learning and Reciprocal Graph.}
\label{tab_gcn_flowers}
\centering
\resizebox{.98\textwidth}{!}{
\begin{tabular}{cll|ccccc}
\hline
\multicolumn{3}{c|}{\textbf{Classifier Specification}} & \multicolumn{5}{c}{\textbf{Feature}}        \\ \hline
 \multirow{2}{*}{\textbf{GCN}} & \multirow{2}{*}{\textbf{Graph}} & \multirow{2}{*}{\textbf{Re-Rank}}  & \textbf{CNN-ResNet} & \textbf{CNN-DPNet} & \textbf{CNN-SENet}  & \textbf{T2T-VIT24}  & \multicolumn{1}{c}{\textbf{VIT-B16}}            \\
& & & \small{\cite{paperCNN_RESNET_2016}} & \small{\cite{paperCNN_DPN2017}} &  \small{\cite{paperCNN_SENET_2018}}  & \small{\cite{paperT2T}}  & \small{\cite{paperVIT16}}            \\
\hline
  \multirow{8}{*}{\rotatebox[origin=c]{70}{GCN-Net}} & kNN & --- & 79.08 $\pm$ 0.3039 & 76.94 $\pm$ 0.3688 & 72.72 $\pm$ 0.2052 & 69.75 $\pm$ 0.0827 & 92.72 $\pm$ 0.1324 \\
  & kNN & LHRR         & 84.37 $\pm$ 0.3239 & 80.76 $\pm$ 0.1372 & 73.89 $\pm$ 0.133  & 72.03 $\pm$ 0.1131 & 95.88 $\pm$ 0.0567 \\
  & kNN & RDPAC        & 83.91 $\pm$ 0.1279 & 81.24 $\pm$ 0.2597 & 74.76 $\pm$ 0.2245 & 74.60 $\pm$ 0.1353  & 96.86 $\pm$ 0.0702 \\
  & kNN & BFSTREE      & 83.12 $\pm$ 0.1784 & 81.39 $\pm$ 0.1222 & 74.83 $\pm$ 0.1284 & 72.49 $\pm$ 0.3283 & 96.33 $\pm$ 0.0695 \\
  & Rec. &  ---             & 83.89 $\pm$ 0.1973 & 81.19 $\pm$ 0.264  & 76.23 $\pm$ 0.1913 & 75.82 $\pm$ 0.2096 & 97.07 $\pm$ 0.0606 \\
  & Rec. & LHRR         & \textbf{84.67 $\pm$ 0.0988} & 80.64 $\pm$ 0.1749 & 73.97 $\pm$ 0.1383 & 72.40 $\pm$ 0.1927  & 95.39 $\pm$ 0.1583 \\
  & Rec. & RDPAC        & 84.20 $\pm$ 0.1975  & \textbf{82.27 $\pm$ 0.1659} & \textbf{76.61 $\pm$ 0.1968} & \textbf{75.87 $\pm$ 0.1877} & \cellcolor{lightgray} \textbf{97.16 $\pm$ 0.0168} \\
  & Rec. &  BFSTREE      & 82.97 $\pm$ 0.1623 & 81.20 $\pm$ 0.1141  & 74.80 $\pm$ 0.2034  & 73.26 $\pm$ 0.1008 & 96.52 $\pm$ 0.0538 \\ \hline
 \multirow{8}{*}{\rotatebox[origin=c]{70}{GCN-SGC}} & kNN  & ---              & 79.64 $\pm$ 0.1023 & 77.09 $\pm$ 0.1139 & 73.00 $\pm$ 0.0941  & 70.05 $\pm$ 0.0802 & 92.84 $\pm$ 0.0655 \\
  & kNN & LHRR         & 84.41 $\pm$ 0.0835 & 80.36 $\pm$ 0.0661 & 74.04 $\pm$ 0.0599 & 72.11 $\pm$ 0.113  & 95.85 $\pm$ 0.0285 \\
  & kNN & RDPAC        & 84.19 $\pm$ 0.0659 & 81.12 $\pm$ 0.0645 & 75.06 $\pm$ 0.0627 & 75.18 $\pm$ 0.0743 & 96.95 $\pm$ 0.0133 \\
  & kNN & BFSTREE      & 83.33 $\pm$ 0.0533 & 81.54 $\pm$ 0.0410  & 75.08 $\pm$ 0.0565 & 72.86 $\pm$ 0.0879 & 96.42 $\pm$ 0.0396 \\
  & Rec.&   ---            & 83.99 $\pm$ 0.0478 & 81.32 $\pm$ 0.0314 & 76.16 $\pm$ 0.0415 & 75.69 $\pm$ 0.0828 & 96.93 $\pm$ 0.0464 \\
  & Rec.& LHRR         & \textbf{84.91 $\pm$ 0.0665} & 80.75 $\pm$ 0.0694 & 74.60 $\pm$ 0.0510   & 72.95 $\pm$ 0.0563 & 95.47 $\pm$ 0.0171 \\
  & Rec.& RDPAC        & 84.53 $\pm$ 0.0580  & \textbf{82.53 $\pm$ 0.1335} & \textbf{76.93 $\pm$ 0.0376} & \textbf{76.43 $\pm$ 0.0499} & \cellcolor{lightgray} \textbf{97.11 $\pm$ 0.0163} \\
  & Rec.& BFSTREE      & 83.43 $\pm$ 0.0200   & 81.58 $\pm$ 0.1169 & 75.03 $\pm$ 0.0313 & 73.58 $\pm$ 0.026  & 96.63 $\pm$ 0.0337 \\ \hline
 \multirow{8}{*}{\rotatebox[origin=c]{70}{GCN-GAT}} & kNN  & ---               & 80.67 $\pm$ 0.2144 & 65.60 $\pm$ 0.9961  & 74.64 $\pm$ 0.3048 & 67.33 $\pm$ 0.9069 & 93.65 $\pm$ 0.228  \\
  & kNN & LHRR         & 84.52 $\pm$ 0.3202 & 76.15 $\pm$ 1.4547 & 75.48 $\pm$ 0.2365 & 73.37 $\pm$ 0.2824 & 95.33 $\pm$ 0.2522 \\
  & kNN & RDPAC        & 84.02 $\pm$ 0.1058 & 77.39 $\pm$ 1.2703 & 75.29 $\pm$ 0.3550 & 75.40 $\pm$ 0.4316  & 97.09 $\pm$ 0.0572 \\
  & kNN & BFSTREE      & 83.04 $\pm$ 0.1844 & 77.19 $\pm$ 1.833  & 75.82 $\pm$ 0.2086 & 73.26 $\pm$ 0.3184 & 96.41 $\pm$ 0.0465 \\
  & Rec.& ---               & 83.67 $\pm$ 0.1965 & 77.42 $\pm$ 0.6762 & 76.74 $\pm$ 0.3398 & 74.82 $\pm$ 0.2978 & 96.99 $\pm$ 0.0558 \\
  & Rec.& LHRR         & \textbf{84.82 $\pm$ 0.2194} & 79.63 $\pm$ 0.6337 & 75.22 $\pm$ 0.2648 & 73.32 $\pm$ 0.3684 & 95.21 $\pm$ 0.2575 \\
  & Rec.& RDPAC        & 84.40 $\pm$ 0.1488  & \textbf{79.69 $\pm$ 1.0373} & \textbf{77.18 $\pm$ 0.2940}  & \textbf{76.90 $\pm$ 0.3418}  & \cellcolor{lightgray} \textbf{97.22 $\pm$ 0.0557} \\
  & Rec.& BFSTREE      & 82.79 $\pm$ 0.2926 & 78.74 $\pm$ 0.2682 & 75.94 $\pm$ 0.2681 & 73.85 $\pm$ 0.2632 & 96.55 $\pm$ 0.0881 \\ \hline
 \multirow{8}{*}{\rotatebox[origin=c]{70}{GCN-APPNP}} & kNN  & ---         & 77.25 $\pm$ 0.1692 & 76.38 $\pm$ 0.238  & 71.0 $\pm$ 0.4051  & 69.45 $\pm$ 0.3072 & 90.24 $\pm$ 0.2128 \\
  & kNN & LHRR    & 84.58 $\pm$ 0.2621 & 82.53 $\pm$ 0.2443 & 76.83 $\pm$ 0.1622 & 74.32 $\pm$ 0.2989 & 96.05 $\pm$ 0.0421 \\
  & kNN & RDPAC   & 85.35 $\pm$ 0.2205 & 83.32 $\pm$ 0.1287 & 76.89 $\pm$ 0.3673 & 77.87 $\pm$ 0.0660  & 97.28 $\pm$ 0.0303 \\
  & kNN & BFSTREE & 84.22 $\pm$ 0.1638 & 83.34 $\pm$ 0.0875 & 77.94 $\pm$ 0.3084 & 75.65 $\pm$ 0.2505 & 96.73 $\pm$ 0.0763 \\
  & Rec.& ---         & 83.91 $\pm$ 0.1181 & 82.20 $\pm$ 0.2160   & 77.74 $\pm$ 0.1645 & 77.11 $\pm$ 0.1485 & 97.24 $\pm$ 0.0470  \\
  & Rec.& LHRR    & \textbf{85.88 $\pm$ 0.1896} & 82.55 $\pm$ 0.2138 & 76.60 $\pm$ 0.2479  & 75.40 $\pm$ 0.2458  & 95.68 $\pm$ 0.1083 \\
  & Rec.& RDPAC   & 85.41 $\pm$ 0.2304 & \textbf{83.99 $\pm$ 0.1276} & \textbf{78.82 $\pm$ 0.1466} & \textbf{78.01 $\pm$ 0.1307} & \cellcolor{lightgray} \textcolor{blue}{\textbf{97.43 $\pm$ 0.0699}} \\
  & Rec.& BFSTREE & 83.75 $\pm$ 0.2099 & 83.14 $\pm$ 0.1915 & 77.83 $\pm$ 0.1826 & 75.85 $\pm$ 0.2098 & 96.89 $\pm$ 0.0632 \\ \hline
 \multirow{8}{*}{\rotatebox[origin=c]{70}{GCN-ARMA}} & kNN  & ---             & 78.69 $\pm$ 0.2471 & 76.01 $\pm$ 0.295  & 73.18 $\pm$ 0.4015 & 70.47 $\pm$ 0.2548 & 91.27 $\pm$ 0.1731 \\
  & kNN & LHRR        & 84.64 $\pm$ 0.3211 & 81.90 $\pm$ 0.4272  & 76.09 $\pm$ 0.1451 & 74.26 $\pm$ 0.2543 & 95.66 $\pm$ 0.1726 \\
  & kNN & RDPAC       & 85.05 $\pm$ 0.1643 & 82.38 $\pm$ 0.3741 & 76.18 $\pm$ 0.3637 & 76.75 $\pm$ 0.2599 & 96.88 $\pm$ 0.0698 \\
  & kNN & BFSTREE     & 83.72 $\pm$ 0.0791 & 81.96 $\pm$ 0.3477 & 76.81 $\pm$ 0.1272 & 75.03 $\pm$ 0.1647 & 96.24 $\pm$ 0.0812 \\
  & Rec.& ---             & 83.32 $\pm$ 0.3713 & 80.86 $\pm$ 0.1282 & 76.96 $\pm$ 0.3041 & 76.11 $\pm$ 0.3851 & 96.66 $\pm$ 0.1140  \\
  & Rec.& LHRR        & \textbf{85.36 $\pm$ 0.3818} & 82.17 $\pm$ 0.3283 & 75.92 $\pm$ 0.2516 & 74.64 $\pm$ 0.3728 & 95.13 $\pm$ 0.2118 \\
  & Rec.& RDPAC       & 84.97 $\pm$ 0.2524 & \textbf{83.14 $\pm$ 0.3078} & \textbf{77.89 $\pm$ 0.2358} & \textbf{77.81 $\pm$ 0.3271} & \cellcolor{lightgray} \textbf{97.02 $\pm$ 0.0944} \\
  & Rec.& BFSTREE     & 84.06 $\pm$ 0.2612 & 82.21 $\pm$ 0.1901 & 76.88 $\pm$ 0.1897 & 75.10 $\pm$ 0.3000     & 96.47 $\pm$ 0.1171 \\
\hline
\end{tabular}
}
\end{table*}

\begin{table*}[!th]
\caption{Impact of manifold learning approaches (LHRR, RDPAC, BFSTREE) and Reciprocal Graph (Rec.) on the classification accuracy (\%) of 5 different GCN models on Corel5k dataset. The best results for each feature and GCN model are highlighted in bold, the best results for each GCN model are marked with a gray background, and the best result for the entire dataset is highlighted in blue. In all the cases, the best results used manifold learning.}
\label{tab_gcn_corel5k}
\centering
\resizebox{.98\textwidth}{!}{
\begin{tabular}{cll|ccccc}
\hline
\multicolumn{3}{c|}{\textbf{Classifier Specification}} & \multicolumn{5}{c}{\textbf{Feature}}        \\ \hline
 \multirow{2}{*}{\textbf{GCN}} & \multirow{2}{*}{\textbf{Graph}} & \multirow{2}{*}{\textbf{Re-Rank}}  & \textbf{CNN-ResNet} & \textbf{CNN-DPNet} & \textbf{CNN-SENet}  & \textbf{T2T-VIT24}  & \multicolumn{1}{c}{\textbf{VIT-B16}}            \\
& & & \small{\cite{paperCNN_RESNET_2016}} & \small{\cite{paperCNN_DPN2017}} &  \small{\cite{paperCNN_SENET_2018}}  & \small{\cite{paperT2T}}  & \small{\cite{paperVIT16}}            \\
\hline
 \multirow{8}{*}{\rotatebox[origin=c]{70}{GCN-Net}}  & kNN &  ---             & 89.34 $\pm$ 0.0950  & 86.49 $\pm$ 0.0998 & 89.17 $\pm$ 0.0956 & 89.02 $\pm$ 0.1452 & 89.93 $\pm$ 0.2878 \\
  & kNN & LHRR         & 91.40 $\pm$ 0.0906  & 88.94 $\pm$ 0.1958 & 90.19 $\pm$ 0.1392 & 90.68 $\pm$ 0.0957 & 94.57 $\pm$ 0.121  \\
  & kNN & RDPAC        & 91.46 $\pm$ 0.1402 & 89.05 $\pm$ 0.1054 & 90.65 $\pm$ 0.0483 & 91.77 $\pm$ 0.1246 & 94.29 $\pm$ 0.139  \\
  & kNN & BFSTREE      & \textbf{92.03 $\pm$ 0.1165} & 89.28 $\pm$ 0.1858 & 91.19 $\pm$ 0.1102 & 91.78 $\pm$ 0.0432 & 94.30 $\pm$ 0.3362  \\
  & Rec. & ---        & 91.68 $\pm$ 0.1064 & \textbf{89.62 $\pm$ 0.1114} & \textbf{91.81 $\pm$ 0.1159} & 92.19 $\pm$ 0.0908 & 93.42 $\pm$ 0.1987 \\
  & Rec. & LHRR         & 91.68 $\pm$ 0.0224 & 88.48 $\pm$ 0.1268 & 90.58 $\pm$ 0.0901 & 91.50 $\pm$ 0.0684  & 94.63 $\pm$ 0.139  \\
  & Rec. & RDPAC        & 92.00 $\pm$ 0.1434  & 89.55 $\pm$ 0.0944 & 90.93 $\pm$ 0.1654 & 91.96 $\pm$ 0.0705 & \cellcolor{lightgray} \textbf{94.76 $\pm$ 0.1577} \\
  & Rec. & BFSTREE      & 92.00 $\pm$ 0.0954  & 89.33 $\pm$ 0.1221 & 91.32 $\pm$ 0.0833 & \textbf{92.43 $\pm$ 0.0401} & 94.39 $\pm$ 0.2771 \\ \hline
 \multirow{8}{*}{\rotatebox[origin=c]{70}{GCN-SGC}} & kNN  & ---              & 89.62 $\pm$ 0.0321 & 86.78 $\pm$ 0.0256 & 89.81 $\pm$ 0.0426 & 88.95 $\pm$ 0.0482 & 93.36 $\pm$ 0.0401 \\
  & kNN & LHRR         & 91.19 $\pm$ 0.0262 & 88.74 $\pm$ 0.0242 & 89.90 $\pm$ 0.044   & 90.49 $\pm$ 0.0518 & 95.20 $\pm$ 0.0219  \\
  & kNN & RDPAC        & 91.47 $\pm$ 0.0216 & 88.95 $\pm$ 0.0632 & 90.70 $\pm$ 0.0403  & 91.77 $\pm$ 0.0521 & 94.76 $\pm$ 0.078  \\
  & kNN & BFSTREE      & 91.98 $\pm$ 0.0246 & 89.23 $\pm$ 0.0453 & 91.40 $\pm$ 0.0061  & 91.71 $\pm$ 0.0444 & 95.26 $\pm$ 0.0759 \\
  & Rec. & ---                & 91.98 $\pm$ 0.0133 & 89.83 $\pm$ 0.0415 & \textbf{92.15 $\pm$ 0.0164} & \textbf{92.75 $\pm$ 0.0908} & 95.49 $\pm$ 0.0107 \\
  & Rec. & LHRR         & 91.73 $\pm$ 0.0508 & 88.70 $\pm$ 0.0669  & 90.73 $\pm$ 0.0235 & 91.68 $\pm$ 0.0305 & \cellcolor{lightgray} \textcolor{blue}{\textbf{95.57 $\pm$ 0.017}}  \\
  & Rec. & RDPAC        & 92.00 $\pm$ 0.0247  & \textbf{89.84 $\pm$ 0.1057} & 90.85 $\pm$ 0.0396 & 92.31 $\pm$ 0.072  & 95.50 $\pm$ 0.020    \\
  & Rec. & BFSTREE      & \textbf{92.04 $\pm$ 0.009}  & 89.49 $\pm$ 0.0627 & 91.30 $\pm$ 0.0257  & 92.54 $\pm$ 0.0591 & 95.30 $\pm$ 0.0479  \\ \hline
 \multirow{8}{*}{\rotatebox[origin=c]{70}{GCN-GAT}} & kNN & ---               & 90.48 $\pm$ 0.1727 & 83.28 $\pm$ 0.33   & 91.13 $\pm$ 0.1107 & 90.7 $\pm$ 0.1187  & 91.3 $\pm$ 0.1764  \\
  & kNN & LHRR         & 92.21 $\pm$ 0.1328 & 88.59 $\pm$ 0.4012 & 91.28 $\pm$ 0.2208 & 92.2 $\pm$ 0.0839  & 94.56 $\pm$ 0.1777 \\
  & kNN & RDPAC        & 91.86 $\pm$ 0.1403 & 89.78 $\pm$ 0.2723 & 91.41 $\pm$ 0.1429 & 92.82 $\pm$ 0.0956 & 94.46 $\pm$ 0.2555 \\
  & kNN & BFSTREE      & \textbf{92.42 $\pm$ 0.1008} & 89.61 $\pm$ 0.362  & 91.95 $\pm$ 0.1382 & 93.09 $\pm$ 0.1337 & 94.58 $\pm$ 0.2226 \\
  & Rec. & ---               & 92.02 $\pm$ 0.0917 & 89.0 $\pm$ 0.2638  & \textbf{92.23 $\pm$ 0.0844} & 92.81 $\pm$ 0.113  & 93.64 $\pm$ 0.2373 \\
  & Rec. & LHRR         & 92.19 $\pm$ 0.1057 & 89.17 $\pm$ 0.2074 & 91.18 $\pm$ 0.1451 & 92.41 $\pm$ 0.1456 & 94.55 $\pm$ 0.1918 \\
  & Rec. & RDPAC        & 92.22 $\pm$ 0.0858 & \textbf{90.48 $\pm$ 0.1718} & 91.48 $\pm$ 0.1021 & 93.02 $\pm$ 0.1334 & \cellcolor{lightgray} \textbf{94.89 $\pm$ 0.1492} \\
  & Rec. & BFSTREE      & 92.30 $\pm$ 0.1128  & 90.01 $\pm$ 0.2374 & 91.88 $\pm$ 0.1081 & \textbf{93.35 $\pm$ 0.1537} & 94.75 $\pm$ 0.1385 \\ \hline
 \multirow{8}{*}{\rotatebox[origin=c]{70}{GCN-APPNP}} & kNN & ---          & 89.72 $\pm$ 0.2031 & 87.68 $\pm$ 0.0785 & 89.92 $\pm$ 0.0992 & 89.86 $\pm$ 0.0731 & 86.89 $\pm$ 0.2487 \\
  & kNN & LHRR    & 92.6 $\pm$ 0.0625  & 90.81 $\pm$ 0.1043 & 91.49 $\pm$ 0.1307 & 91.83 $\pm$ 0.0952 & 94.53 $\pm$ 0.1144 \\
  & kNN & RDPAC   & 92.69 $\pm$ 0.1161 & 90.75 $\pm$ 0.179  & 91.81 $\pm$ 0.098  & 92.58 $\pm$ 0.0588 & 94.12 $\pm$ 0.2213 \\
  & kNN & BFSTREE & 93.04 $\pm$ 0.0872 & \textbf{91.01 $\pm$ 0.1026} & 92.35 $\pm$ 0.0535 & 92.83 $\pm$ 0.031  & 94.37 $\pm$ 0.0855 \\
  & Rec. & ---           & 92.69 $\pm$ 0.05   & 90.70 $\pm$ 0.1301  & \textbf{92.79 $\pm$ 0.0429} & 93.56 $\pm$ 0.0669 & 93.53 $\pm$ 0.1042 \\
  & Rec. & LHRR    & 92.88 $\pm$ 0.1058 & 89.99 $\pm$ 0.0869 & 91.78 $\pm$ 0.0694 & 92.63 $\pm$ 0.0817 & 94.95 $\pm$ 0.2116 \\
  & Rec. & RDPAC   & 92.82 $\pm$ 0.046  & 90.95 $\pm$ 0.1134 & 91.92 $\pm$ 0.0738 & 93.17 $\pm$ 0.0804 & \cellcolor{lightgray} \textbf{95.13 $\pm$ 0.1095} \\
  & Rec. & BFSTREE & \textbf{93.08 $\pm$ 0.0727} & 90.78 $\pm$ 0.1317 & 92.39 $\pm$ 0.0269 & \textbf{93.70 $\pm$ 0.0653}  & 94.72 $\pm$ 0.1564 \\ \hline
 \multirow{8}{*}{\rotatebox[origin=c]{70}{GCN-ARMA}} & kNN & ---               & 88.58 $\pm$ 0.312  & 86.47 $\pm$ 0.0729 & 89.11 $\pm$ 0.1061 & 89.16 $\pm$ 0.0571 & 85.48 $\pm$ 0.3945 \\
  & kNN & LHRR        & 91.58 $\pm$ 0.1185 & 89.84 $\pm$ 0.1565 & 90.98 $\pm$ 0.1738 & 91.46 $\pm$ 0.0963 & 90.66 $\pm$ 0.5051 \\
  & kNN & RDPAC       & 91.72 $\pm$ 0.1775 & 90.09 $\pm$ 0.2858 & 91.08 $\pm$ 0.1226 & 92.28 $\pm$ 0.0545 & 92.62 $\pm$ 0.4067 \\
  & kNN & BFSTREE     & 92.23 $\pm$ 0.1447 & 90.31 $\pm$ 0.1195 & 91.7 $\pm$ 0.0869  & 92.24 $\pm$ 0.0682 & 92.28 $\pm$ 0.3061 \\
  & Rec. & ---              & 91.14 $\pm$ 0.137  & 89.24 $\pm$ 0.2139 & 91.31 $\pm$ 0.1887 & 91.84 $\pm$ 0.0774 & 90.48 $\pm$ 0.1707 \\
  & Rec. & LHRR        & 91.77 $\pm$ 0.1541 & 89.24 $\pm$ 0.1428 & 91.07 $\pm$ 0.126  & 91.78 $\pm$ 0.1145 & 92.39 $\pm$ 0.2078 \\
  & Rec. & RDPAC       & 92.05 $\pm$ 0.1403 & \textbf{90.41 $\pm$ 0.1645} & 91.47 $\pm$ 0.1202 & 92.49 $\pm$ 0.2056 & 92.80 $\pm$ 0.1896  \\
  & Rec. & BFSTREE     & \textbf{92.27 $\pm$ 0.0377} & 90.14 $\pm$ 0.1897 & \textbf{91.71 $\pm$ 0.1753} & \cellcolor{lightgray} \textbf{92.90 $\pm$ 0.1446}  & \textbf{92.74 $\pm$ 0.2083} \\
\hline
\end{tabular}
}
\end{table*}

\begin{table*}[!t]
\caption{Impact of manifold learning approaches (LHRR, RDPAC, BFSTREE) and Reciprocal Graph (Rec.) on the classification accuracy (\%) of 5 different GCN models on CUB200 dataset. The best results for each feature and GCN model are highlighted in bold, the best results for each GCN model are marked with a gray background, and the best result for the entire dataset is highlighted in blue. In all the cases, the best results used manifold learning.}
\label{tab_gcn_cub200}
\centering
\resizebox{.90\textwidth}{!}{
\begin{tabular}{cll|cccc}
\hline
\multicolumn{3}{c|}{\textbf{Classifier Specification}} & \multicolumn{4}{c}{\textbf{Feature}}        \\ \hline
 \multirow{2}{*}{\textbf{GCN}} & \multirow{2}{*}{\textbf{Graph}} & \multirow{2}{*}{\textbf{Re-Rank}}  & \textbf{CNN-ResNet} & \textbf{CNN-SENet} & \textbf{CNN-Xception} & \multicolumn{1}{c}{\textbf{VIT-B16}}            \\
 & & & \small{\cite{paperCNN_RESNET_2016}} & \small{\cite{paperCNN_DPN2017}} &  \small{\cite{paperCNN_XCEPTION_2017}}  & \small{\cite{paperVIT16}}            \\ \hline
 \multirow{8}{*}{\rotatebox[origin=c]{70}{GCN-Net}} & kNN & ---                 & 40.76 $\pm$ 0.7467 & 35.8 $\pm$ 0.0634  & 46.66 $\pm$ 0.019  & 64.39 $\pm$ 0.4486 \\
  & kNN & LHRR         & 49.16 $\pm$ 0.3119 & 36.17 $\pm$ 0.1153 & 51.13 $\pm$ 0.0738 & 70.42 $\pm$ 0.671  \\
  & kNN & RDPAC        & 49.44 $\pm$ 0.1092 & 36.84 $\pm$ 0.0578 & 51.18 $\pm$ 0.0284 & 72.71 $\pm$ 0.1506 \\
  & kNN & BFSTREE      & 49.18 $\pm$ 0.1011 & 37.10 $\pm$ 0.0482  & 50.62 $\pm$ 0.0639 & 71.54 $\pm$ 0.1888 \\
  & Rec. & ---         & 49.46 $\pm$ 0.3279 & \textbf{39.42 $\pm$ 0.113}  & 50.76 $\pm$ 0.0713 & 68.85 $\pm$ 0.3055 \\
  & Rec. & LHRR         & 51.23 $\pm$ 0.0788 & 36.5 $\pm$ 0.0728  & 51.92 $\pm$ 0.0546 & 73.49 $\pm$ 0.1879 \\
  & Rec. & RDPAC        & \textbf{51.57 $\pm$ 0.0999} & 38.57 $\pm$ 0.0712 & \textbf{53.12 $\pm$ 0.0596} & \cellcolor{lightgray}  \textbf{74.39 $\pm$ 0.3061} \\
  & Rec. & BFSTREE      & 50.80 $\pm$ 0.0291  & 37.8 $\pm$ 0.0538  & 51.82 $\pm$ 0.0658 & 73.58 $\pm$ 0.3939 \\ \hline
 \multirow{8}{*}{\rotatebox[origin=c]{70}{GCN-SGC}} & kNN & ---                 & 47.55 $\pm$ 0.0329 & 36.48 $\pm$ 0.0684 & 48.60 $\pm$ 0.0072  & 74.23 $\pm$ 0.0385 \\
  & kNN & LHRR         & 51.22 $\pm$ 0.0184 & 35.88 $\pm$ 0.0137 & 52.36 $\pm$ 0.0125 & 77.84 $\pm$ 0.0519 \\
  & kNN & RDPAC        & 51.88 $\pm$ 0.0315 & 37.75 $\pm$ 0.0148 & 52.98 $\pm$ 0.0103 & 78.16 $\pm$ 0.0453 \\
  & kNN & BFSTREE      & 51.66 $\pm$ 0.016  & 37.70 $\pm$ 0.01    & 52.21 $\pm$ 0.0095 & 77.31 $\pm$ 0.0563 \\
  & Rec. & ---                 & 53.71 $\pm$ 0.0362 & \textbf{40.31 $\pm$ 0.0255} & 54.0 $\pm$ 0.0054  & 78.03 $\pm$ 0.0428 \\
  & Rec. & LHRR         & 51.99 $\pm$ 0.0251 & 36.74 $\pm$ 0.0162 & 53.12 $\pm$ 0.0153 & 78.54 $\pm$ 0.0177 \\
  & Rec. & RDPAC        & \textbf{52.85 $\pm$ 0.0164} & 38.91 $\pm$ 0.0073 & \textbf{54.59 $\pm$ 0.0036} & \cellcolor{lightgray}  \textcolor{blue}{\textbf{79.27 $\pm$ 0.0325}} \\
  & Rec. & BFSTREE      & 52.68 $\pm$ 0.0308 & 38.65 $\pm$ 0.023  & 53.54 $\pm$ 0.0041 & 78.12 $\pm$ 0.0344 \\ \hline
 \multirow{8}{*}{\rotatebox[origin=c]{70}{GCN-GAT}} & kNN & ---                 & 41.84 $\pm$ 0.2901 & 32.5 $\pm$ 0.205   & 42.45 $\pm$ 0.1848 & 59.53 $\pm$ 0.5668 \\
  & kNN & LHRR         & 48.86 $\pm$ 0.1593 & 34.78 $\pm$ 0.1155 & 48.8 $\pm$ 0.246   & 64.02 $\pm$ 0.4082 \\
  & kNN & RDPAC        & 49.05 $\pm$ 0.1145 & 35.9 $\pm$ 0.1158  & 49.03 $\pm$ 0.1037 & 68.78 $\pm$ 0.2495 \\
  & kNN & BFSTREE      & 48.77 $\pm$ 0.1427 & 35.98 $\pm$ 0.1457 & 48.3 $\pm$ 0.1084  & 68.1 $\pm$ 0.2488  \\
  & Rec. & ---                 & 45.46 $\pm$ 0.1879 & 33.02 $\pm$ 0.1206 & 45.88 $\pm$ 0.16   & 64.82 $\pm$ 0.2582 \\
  & Rec. & LHRR         & 50.19 $\pm$ 0.0904 & 35.28 $\pm$ 0.1364 & 50.17 $\pm$ 0.1073 & 70.31 $\pm$ 0.0762 \\
  & Rec. & RDPAC        & \textbf{50.95 $\pm$ 0.0632} & \textbf{37.55 $\pm$ 0.1087} & \textbf{51.29 $\pm$ 0.1577} & \cellcolor{lightgray}  \textbf{72.94 $\pm$ 0.1716} \\
  & Rec. & BFSTREE      & 49.89 $\pm$ 0.1871 & 36.67 $\pm$ 0.129  & 49.87 $\pm$ 0.1245 & 71.73 $\pm$ 0.1775 \\ \hline
 \multirow{8}{*}{\rotatebox[origin=c]{70}{GCN-APPNP}} & kNN & ---           & 29.16 $\pm$ 0.6867 & 30.27 $\pm$ 0.3694 & 42.68 $\pm$ 0.0826 & 55.24 $\pm$ 0.5689 \\
  & kNN & LHRR    & 47.0 $\pm$ 0.1836  & 34.91 $\pm$ 0.1598 & 48.77 $\pm$ 0.0979 & 66.57 $\pm$ 0.572  \\
  & kNN & RDPAC   & 47.19 $\pm$ 0.0701 & 35.29 $\pm$ 0.1195 & 47.72 $\pm$ 0.094  & 69.92 $\pm$ 0.2262 \\
  & kNN & BFSTREE & 46.59 $\pm$ 0.2154 & 35.28 $\pm$ 0.0718 & 47.14 $\pm$ 0.0895 & 70.86 $\pm$ 0.2702 \\
  & Rec. & ---           & 48.51 $\pm$ 0.1192 & 38.02 $\pm$ 0.0461 & 47.51 $\pm$ 0.0452 & 68.29 $\pm$ 0.0935 \\
  & Rec. & LHRR    & \textbf{51.99 $\pm$ 0.0800}   & 37.45 $\pm$ 0.0768 & 51.43 $\pm$ 0.084  & 74.61 $\pm$ 0.0991 \\
  & Rec. & RDPAC   & 51.82 $\pm$ 0.1028 & \textbf{39.15 $\pm$ 0.1601} & \textbf{52.17 $\pm$ 0.0865} & \cellcolor{lightgray}  \textbf{75.59 $\pm$ 0.2139} \\
  & Rec. & BFSTREE & 50.6 $\pm$ 0.0848  & 38.21 $\pm$ 0.0358 & 50.26 $\pm$ 0.1301 & 74.15 $\pm$ 0.1837 \\ \hline
 \multirow{8}{*}{\rotatebox[origin=c]{70}{GCN-ARMA}} & kNN & ---              & 38.74 $\pm$ 0.4527 & 32.96 $\pm$ 0.1626 & 42.91 $\pm$ 0.1465 & 60.26 $\pm$ 0.4398 \\
  & kNN & LHRR        & 47.58 $\pm$ 0.2387 & 34.56 $\pm$ 0.0799 & 49.26 $\pm$ 0.2191 & 67.21 $\pm$ 0.2825 \\
  & kNN & RDPAC       & 47.77 $\pm$ 0.2075 & 35.4 $\pm$ 0.1474  & 49.88 $\pm$ 0.1479 & 71.16 $\pm$ 0.2337 \\
  & kNN & BFSTREE     & 47.12 $\pm$ 0.3126 & 35.6 $\pm$ 0.1385  & 48.78 $\pm$ 0.0991 & 70.13 $\pm$ 0.4433 \\
  & Rec. & ---               & 44.37 $\pm$ 0.1739 & 34.25 $\pm$ 0.1559 & 46.95 $\pm$ 0.3062 & 64.55 $\pm$ 0.3184 \\
  & Rec. & LHRR        & 49.29 $\pm$ 0.0987 & 35.22 $\pm$ 0.0891 & 50.38 $\pm$ 0.1318 & 70.05 $\pm$ 0.653  \\
  & Rec. & RDPAC       & \textbf{49.81 $\pm$ 0.2090}  & \textbf{37.12 $\pm$ 0.1276} & \textbf{51.63 $\pm$ 0.1155} & \cellcolor{lightgray}  \textbf{73.29 $\pm$ 0.34}   \\
  & Rec. & BFSTREE     & 48.92 $\pm$ 0.2721 & 36.38 $\pm$ 0.1331 & 50.41 $\pm$ 0.0777 & 72.17 $\pm$ 0.3336 \\
\hline
\end{tabular}
}
\end{table*}

%%%% TSNE %%%%

%----------------------------------------------------------------
\subsection{Comparison with Other Approaches}
\label{ssec:comparison}
%----------------------------------------------------------------

For comparison purposes, a wide variety of supervised and semi-supervised classification approaches were considered, both traditional and more recent ones.
A brief description of the employed baselines, the implementations, and the parameters used are presented in the following:

\begin{itemize}
    \item \textbf{$k$ Nearest Neighbor (kNN)}: traditional approach that computes the distance to the other elements in the dataset and selects the $k$ closest ones. The sklearn implementation was used, with $k=20$.
    \item \textbf{Support Vector Machine (SVM)}~\cite{paperSVM}:
    it is a traditional method that consists in finding the hyperplane that best separates the data into the correct classes in a high-dimensional space.
    The sklearn implementation was used, with default parameters and Radial Basis Function (RBF) kernel.
    \item \textbf{Single-Layer and Multi-Layer Perceptron}:
    the sklearn implementation was used for both, with the Stochastic Gradient Descent (SGD) optimizer.
    \item \textbf{Optimum-Path Forest (OPF)}~\cite{PaperOPFSupervised2009, paperSSOPF2014}:
    it builds a graph where each node is an element of the dataset and the edges are weighted by their Euclidean distance. The algorithm computes the optimum path between the nodes in order to classify them into a given class.
    The pyOPF~\footnote{\url{https://github.com/marcoscleison/PyOPF}} implementation was used, with the default parameters.
    \item \textbf{Pseudo-label}~\cite{lee2013pseudo}:
     the method is semi-supervised and is used to assign labels to unlabeled data.
    In this work, a public implementation~\footnote{\url{https://github.com/anirudhshenoy/pseudo_labeling_small_datasets}} was used along with the Logistic Regression classifier that employed the Stochastic Gradient Descent (SGD) optimizer and Squared Hinge loss with $\alpha = 10^{-5}$ for training.
    \item \textbf{Label Spreading (LS)}~\cite{Zhou04learningwith}: a semi-supervised algorithm that attributes labels to elements according to the labels of their neighbors, given a certain degree of similarity. For this process, it uses an affinity matrix based on a normalized graph Laplacian.
    The sklearn implementation was used, considering a Radial Basis Function (RBF) kernel with $\alpha = 0.4125$, $\gamma = 0.1$, and a maximum of 100 iterations. This method is used to expand the training set and is used along with the other classifiers.
    \item \textbf{Learning Discrete Structures for Graph Neural Networks (GNN-LDS and GNN-KNN-LDS)}~\cite{paperLDS_GNN_2019}: 
    this Graph Neural Network (GNN) learns both a graph and embeddings from the input features. It approximately solves a bilevel program that learns a discrete probability distribution on the edges of the graph.
    The authors claim that this is the first method that simultaneously learns the graph and the parameters of a GNN for semi-supervised classification.
    The approach presents two variants: (\emph{i}) GNN-LDS; and (\emph{ii}) GNN-KNN-LDS which initializes by computing a kNN graph.
    Both were used as baselines with their default parameters proposed in the implementation~\footnote{\url{https://github.com/lucfra/LDS-GNN}} provided by the original authors. For the kNN graph, $k=20$ was used.
    
    \item \textbf{Weakly Supervised Framework Experiments Framework (WSEF)}~\cite{paperWSJoao2020}: 
    the method generates pseudo-labels by applying different rank correlation measures (e.g., Jaccard, Spearman). The approach is mainly based on the idea that elements that have ranked lists with a high intersection with others probably belong to the same class.
    The implementation~\footnote{\url{https://github.com/UDLF/WSEF}} provided by the authors was used considering Rank Biased Overlap (RBO)~\cite{paperRBO} correlation measure, $k=40$ in combination with SVM.
    \item \textbf{CoMatch}~\cite{paperCoMatch}: the method is based on concepts of graph-based self-supervised learning. The approach is trained to produce similar embeddings for the same image with different augmentations. CoMatch jointly optimizes three losses: (\emph{i}) a supervised classification loss on labeled data, (\emph{ii}) an unsupervised classification loss on unlabeled data, and (\emph{iii}) a graph-based contrastive loss on unlabeled data.
    It takes images as input instead of features. This version employs ResNet~\cite{paperCNN_RESNET_2016} as the backbone.
    We considered the implementation provided by the authors~\footnote{\url{https://github.com/salesforce/CoMatch}}, with default parameters (the ones used for ImageNet~\cite{paperImagenet2009} in their code). We trained with a batch size of 25 and 400 epochs for all datasets. Except for the CUB200 dataset, which is larger, we used a batch size of 50 and 300 epochs.
\end{itemize}

We also compared our results with three CNN-based classifiers, considering image data as input.
The images were provided with a size of 100x100 pixels in batches of size 32.
For the other methods, the input consists of feature vectors obtained from deep features trained through transfer learning.
Notice that the CNNs used as baselines require more labeled data in comparison to other methods and were evaluated on a supervised cross-validation scenario  (9 folds for training, 1 fold for testing).
Except for CNN classifiers, all other methods were evaluated on semi-supervised scenarios  (1 fold for training, 9 folds for testing).

Table~\ref{tab:baselines_classification} presents the comparison with both traditional and recent state-of-the-art baselines in relation to our approach on Flowers, Corel5k, and CUB200 datasets.
Most results are the mean of 5 executions of 10 folds, with some exceptions which are indicated in italic text.
Some methods require long running times on larger datasets (i.e., LDS and CoMatch).
For GNN-KNN-LDS, KNN-LDS, and CoMatch, the results on CUB200 correspond to 1 execution.
For CoMatch, the mean of 3 executions is reported for Corel5k dataset.
The best result for each feature is highlighted in bold and the best for each dataset is highlighted in red.
The gray rows indicate the results that correspond to our method.

The proposed method revealed superior results compared to the baselines in most of the cases.
The only exception is Flowers with VIT-B16 features where WSEF shows the best results (97.82\% accuracy).
However, our ManifoldGCN is very close with 97.43\% accuracy.

% k=20 used for lds and knn, manifoldgcn uses k=40
% lds uses pca to reduce to 100 features
\begin{table*}[!t]
%\vspace{-35mm}
\centering
\caption{Accuracy comparison (\%) for baselines on Flowers, Corel5k, and CUB200 datasets. For every dataset, we compared our approach with both supervised and semi-supervised baselines. The methods are compared with different input features. The results of our method are highlighted with a gray background; the best results for each pair of features and dataset are marked in bold, and the best results for each dataset are in red.}
\label{tab:baselines_classification}
\resizebox{.9\textwidth}{!}{ 
\begin{tabular}{l|c|l|c|ccc}
\hline
\textbf{Method} & \textbf{Year} & \textbf{Input} & \textbf{Training} & \textbf{Flowers} & \textbf{Corel5k} & \textbf{CUB200} \\ \hline \hline
\textbf{MobileNet} & 2017 &  & \multirow{ 3}{*}{Supervised} & 86.66 & 90.90 & 35.20 \\
\textbf{ResNet50} & 2015 & ~\textbf{Images} & & 85.97 & 91.52  & 31.10 \\
\textbf{CNN-Xception} & 2016 & & & 90.24 & 93.32 & 44.25  \\
\hline \hline
\textbf{CoMatch} & 2021 & ~\textbf{Images} & \multirow{ 1}{*}{Semi-Supervised} &  82.55 & \emph{85.70} & \emph{38.29} \\
\hline
\textbf{kNN} & --- & & \multirow{ 16}{*}{Semi-Supervised} & 63.67 & 76.80 & 36.67 \\
\textbf{SVM} & 1995 &  & & 80.54 & 88.73 & 48.84 \\
\textbf{OPF} & 2009 & & & 71.77 & 83.56 & 38.59 \\
\textbf{SL-Perceptron} & ---  & & & 75.44 & 83.56 & 39.91 \\
\textbf{ML-Perceptron} & --- & & & 78.88 & 87.10 & 32.24 \\
\textbf{PseudoLabel+SGD} & 2013 & & & 82.69 & 89.76 & 21.67  \\
\textbf{LS+kNN} & 2004 & \textbf{ResNet} & & 73.49 & 83.98 & 36.99  \\
\textbf{LS+SVM} & 2004 & \textbf{Features} & & 73.53 & 83.26  & 38.70   \\
\textbf{LS+OPF} & 2004 & & & 72.66 & 82.32  & 39.28 \\
\textbf{LS+SL-Perceptron} & 2004 & & & 72.34 & 82.38 & 39.21    \\
\textbf{LS+ML-Perceptron} & 2004 & & & 73.03 & 82.53 & 39.68   \\
\textbf{GNN-LDS} & 2019 & & & 54.98 & 62.69 & ---  \\
\textbf{GNN-KNN-LDS} & 2019 &  & & 79.32 & 88.94 & \emph{37.78}  \\
\textbf{WSEF+SVM+RBO} & 2021 & & & 85.12  & 91.68 & 52.17  \\
\cellcolor{lightgray} \textbf{SGC+Rec.+RDPAC} & Ours & & & \cellcolor{lightgray}  84.53 &  \cellcolor{lightgray}  92.00 &  \cellcolor{lightgray}  \textbf{52.85} \\
\cellcolor{lightgray} \textcolor{blue}{\textbf{ManifoldGCN (best result)}} & Ours &  & & \cellcolor{lightgray}  \textbf{85.88} &  \cellcolor{lightgray}  \textbf{93.08} &  \cellcolor{lightgray}  \textbf{52.85} \\
%\cline{4-6}
\hline

\textbf{kNN} & --- &  & \multirow{ 16}{*}{Semi-Supervised} & 48.71 & 58.78 & 22.23 \\
\textbf{SVM} & 1995 & & & 73.30 & 85.89 & 35.32 \\
\textbf{OPF} & 2009 & & & 64.00 & 81.33 & 30.94 \\
\textbf{SL-Perceptron} & --- & & & 71.84 & 82.28 & 36.39 \\
\textbf{ML-Perceptron} & --- & & & 72.62 & 86.90 & 32.15 \\
\textbf{PseudoLabel+SGD} & 2013 & & & 76.87 & 89.85 & 20.96  \\
\textbf{LS+kNN} & 2004 & \textbf{SENet} & & 58.05 & 72.16 & 20.00   \\
\textbf{LS+SVM} & 2004 & \textbf{Features} & & 59.84  & 72.79 & 24.82  \\
\textbf{LS+OPF} & 2004 & & & 59.25 & 72.20  & 25.38  \\
\textbf{LS+SL-Perceptron} & 2004 &  & & 59.27 & 72.19 & 25.41   \\
\textbf{LS+ML-Perceptron} & 2004 & & & 59.39  & 72.24  & 25.72  \\
\textbf{GNN-LDS} & 2019 & & & 52.24 &  65.80 & ---  \\
\textbf{GNN-KNN-LDS} & 2019 &  & & 73.69 & 89.95 & ---  \\
\textbf{WSEF+SVM+RBO} & 2021 &  & & 76.16  & 89.74 & 36.49  \\
\cellcolor{lightgray} \textbf{SGC+Rec.+RDPAC} & Ours & & & \cellcolor{lightgray}  76.93 &  \cellcolor{lightgray}  90.85 &  \cellcolor{lightgray}  38.91 \\
\cellcolor{lightgray} \textcolor{blue}{\textbf{ManifoldGCN (best result)}} &  Ours &  & & \cellcolor{lightgray}  \textbf{78.82} &  \cellcolor{lightgray}  \textbf{92.79} &  \cellcolor{lightgray}  \textbf{40.31} \\
\hline

\textbf{kNN} & --- &  &  \multirow{ 16}{*}{Semi-Supervised} & 91.91 & 81.19 &  56.62 \\
\textbf{SVM} & 1995 &  & & 96.75 & 91.92 &  75.61 \\
\textbf{OPF} & 2009 & & & 96.50 & 90.02 &  73.27 \\
\textbf{SL-Perceptron} & --- & & & 75.79 & 82.15 & 70.84 \\
\textbf{ML-Perceptron} & --- & & & 92.59 & 74.41 &  12.02 \\
\textbf{PseudoLabel+SGD} & 2013 & & & 96.84 & 89.07  & 30.19 \\
\textbf{LS+kNN} & 2004 & \textbf{VIT-B16} & & 95.74 & 89.63 & 66.15   \\
\textbf{LS+SVM} & 2004 & \textbf{Features} & & 94.49 & 87.59 & 66.81  \\
\textbf{LS+OPF} & 2004 & & & 94.22  & 86.14  & 66.68 \\
\textbf{LS+SL-Perceptron} & 2004 & & & 93.71 & 86.31 & 65.45   \\
\textbf{LS+ML-Perceptron} & 2004 & & & 95.13 & 87.68 & 62.81  \\
\textbf{GNN-LDS} & 2019 &  & & 72.03 & 56.33 & \emph{22.75}   \\
\textbf{GNN-KNN-LDS} & 2019 & & & 96.66 & 88.56  & \emph{52.42}   \\
\textbf{WSEF+SVM+RBO} & 2021 & & & \textcolor{red}{\textbf{97.82}}  & 94.00 & 78.64  \\
\cellcolor{lightgray} \textbf{SGC+Rec.+RDPAC} & Ours & & & \cellcolor{lightgray}  97.11 &  \cellcolor{lightgray}  95.50 &  \cellcolor{lightgray}  \textbf{79.27} \\
\cellcolor{lightgray} \textcolor{blue}{\textbf{ManifoldGCN (best result)}} & Ours  & & & \cellcolor{lightgray}  97.43 &  \cellcolor{lightgray}  \textcolor{red}{\textbf{95.57}} &  \cellcolor{lightgray}  \textcolor{red}{\textbf{79.27}} \\ \hline
\end{tabular}
}
\end{table*}

\subsection{Efficiency Results}
\label{ssec:effeciencyresults}

We conducted an experiment to measure the run-time (in seconds) for running each of the manifold learning methods and GCN models.
The experiments were executed on a machine with an
Intel(R) Core(TM) i7-10700F CPU @ 2.90GHz, 32 GB RAM, NVIDIA GeForce RTX 3060 GPU with 12GB VRAM running Ubuntu 20.04 with Linux kernel 5.15.0-52-generic.
Table~\ref{tab:runtime} reports the average and standard deviation of 5 executions of 10 folds on each dataset and for the two types of graphs ($kNN$ and Reciprocal $kNN$).

Manifold Learning (M.L.) performs the pre-processing of the GCN graph. Since these methods are not currently parallelized, they run all on CPU. Parallelization of these approaches is out of the scope of this paper, but rank-based methods can be parallelized with data parallelism as shown in other papers~\cite{paperRLRecom, paperCPRR_PRL2017}.
While the training involves both the GCN initialization and the learning process, testing is responsible for computing the classification of all the queries. Both training and testing are performed on the GPU.

Notice that the execution times are very low, which indicates that the method is fast even for the more robust GCNs.
Also, most compared methods have a costly training process.
An example is CoMatch (2021) that requires huge training times: 40 minutes on Flowers; 88.4 minutes on Corel5k; 260 minutes on CUB200.
These times are the average of executions for 100 epochs.
However, CoMatch is generally recommended to be trained for 400 epochs.
These values are much higher than our proposed approach.

\begin{table}[!ht]
\centering
\caption{Execution time (in seconds) for manifold learning methods and GCN approaches for both training and testing.}
\label{tab:runtime}
\resizebox{.49\textwidth}{!}{ 
\begin{tabular}{lllll}
\hline
      &                  & \textbf{Flowers}        & \textbf{Corel5k} & \textbf{CUB200} \\
\hline
\textbf{\multirow{3}{*}{\rotatebox{90}{M.L.}}}      & LHRR             & 1.10 $\pm$ 0.0012 & 6.21 $\pm$ 0.0017        & 20.16 $\pm$ 0.0108       \\
  & RDPAC            & 4.66 $\pm$ 0.0195 & 41.74 $\pm$ 0.0158        & 104.18 $\pm$ 0.5091       \\
      & BFSTREE          & 9.34 $\pm$ 0.0046 & 37.94 $\pm$ 0.1712        & 95.09 $\pm$ 0.0704       \\
\hline
\textbf{\multirow{10}{*}{\rotatebox{90}{Train}}} & GCN-Net (kNN)    & 0.76 $\pm$ 0.0187 & 2.23 $\pm$ 0.0178    & 6.95 $\pm$ 0.0141       \\
      & GCN-Net (Rec.)   & 0.61 $\pm$ 0.0016 & 1.57 $\pm$ 0.0018    & 4.40 $\pm$ 0.0003       \\
      & GCN-SGC (kNN)    & 0.15 $\pm$ 0.0005 & 0.20 $\pm$ 0.0011    & 0.54 $\pm$ 0.0006      \\
      & GCN-SGC (Rec.)   & 0.14 $\pm$ 0.0003 & 0.19 $\pm$ 0.0022    & 0.51 $\pm$ 0.0002      \\
      & GCN-GAT (kNN)    & 3.41 $\pm$ 0.0021 & 11.89 $\pm$ 0.0031   & 30.62 $\pm$ 0.0095       \\
      & GCN-GAT (Rec.)   & 2.44 $\pm$ 0.0025 & 7.90 $\pm$ 0.0029    & 19.35 $\pm$ 0.0043        \\
      & GCN-APPNP (kNN)  & 0.77 $\pm$ 0.0088 & 3.49 $\pm$ 0.0032    & 27.95 $\pm$ 0.0025        \\
      & GCN-APPNP (Rec.) & 0.75 $\pm$ 0.0002 & 2.44 $\pm$ 0.0032    & 17.11 $\pm$ 0.0032        \\
      & GCN-ARMA (kNN)   & 3.84 $\pm$ 0.0064 & 14.45 $\pm$ 0.0215   & 47.8 $\pm$ 0.0146      \\
      & GCN-ARMA (Rec.)  & 2.80 $\pm$ 0.0051 & 9.94 $\pm$ 0.0013    & 30.51 $\pm$ 0.0113      \\
\hline
\textbf{\multirow{10}{*}{\rotatebox{90}{Test}}}  & GCN-Net (kNN)    & 0.06 $\pm$ 0.0366 & 0.18 $\pm$ 0.0382    & 0.40 $\pm$ 0.0327      \\
      & GCN-Net (Rec.)   & 0.05 $\pm$ 0.0013 & 0.18 $\pm$ 0.002     & 0.44 $\pm$ 0.0029       \\
      & GCN-SGC (kNN)    & 0.04 $\pm$ 0.0015 & 0.16 $\pm$ 0.0011    & 0.38 $\pm$ 0.0051       \\
      & GCN-SGC (Rec.)   & 0.05 $\pm$ 0.0015 & 0.18 $\pm$ 0.0018    & 0.44 $\pm$ 0.0005        \\
      & GCN-GAT (kNN)    & 0.04 $\pm$ 0.001  & 0.15 $\pm$ 0.0009    & 0.38 $\pm$ 0.004       \\
      & GCN-GAT (Rec.)   & 0.05 $\pm$ 0.0015 & 0.18 $\pm$ 0.002     & 0.44 $\pm$ 0.0032       \\
      & GCN-APPNP (kNN)  & 0.05 $\pm$ 0.0015 & 0.15 $\pm$ 0.0008    & 0.38 $\pm$ 0.0045       \\
      & GCN-APPNP (Rec.) & 0.05 $\pm$ 0.0015 & 0.18 $\pm$ 0.0021    & 0.44 $\pm$ 0.0026    \\
      & GCN-ARMA (kNN)   & 0.04 $\pm$ 0.0015 & 0.15 $\pm$ 0.0008    & 0.39 $\pm$ 0.0036       \\
      & GCN-ARMA (Rec.)  & 0.04 $\pm$ 0.0018 & 0.18 $\pm$ 0.0023    & 0.44 $\pm$ 0.0004      \\
\hline
\end{tabular}
}
\end{table}

\section{Conclusions}
\label{sec:conclusion}

In this work, we presented a novel framework, the Manifold-GCN, for semi-supervised image classification.
The approach is flexible and allows the use of different types of GCNs, graphs, features, and manifold learning.
To the best of our knowledge, this is the first framework that allows the combination of GCNs with different types of manifold learning approaches for image classification. In our work, all the manifold learning algorithms are completely unsupervised, which is especially useful for scenarios where labeled data is limited.

An experimental evaluation was conducted on 3 datasets and 5 distinct deep learning features, including features obtained by CNNs and Visual Transformers.
The results validated our main hypothesis by revealing that the manifold learning methods were capable of improving the effectiveness (accuracy and F-Measure) in the vast majority of the cases, for different GCNs and features.
The visual analyzes show that the feature space was significantly improved by the proposed approach.
In comparison to both traditional and very recent state-of-the-art methods, it is clearly noticeable that Manifold-GCN is among the best in most cases.
Our approach is also very efficient and requires very low execution times for training and testing.

In future work, we intend to use other types of graph models in addition to kNN Graph and Reciprocal Graph.
For this work, the input feature and the feature used to model the graph are the same. Since complementary features can possibly increase results even further, we intend to evaluate cases where a different feature can be used for the graph.
We also intend to employ Manifold-GCN in larger image collections and retrieval scenarios.
For efficiency purposes, it is also possible to parallelize the implementations of manifold learning approaches.

%\vspace{-2mm}
%***************************************************************************************
\section*{\uppercase{Acknowledgments}}
%***************************************************************************************
The authors are grateful to Fulbright Commission, São Paulo Research Foundation - FAPESP (grants \#2017/25908-6, \#2018/15597-6, and \#2020/11366-0), Brazilian National Council for Scientific and Technological Development - CNPq (grant \#309439/2020-5), Petrobras (grant \#2017/00285-6), and Microsoft Research for financial support.
We also acknowledge the NSF Grant No. IIS-2107213.

\ifCLASSOPTIONcaptionsoff
  \newpage
\fi




%***************************************** BIBLIOGRAPHY ************************************************
%\small
%\def\baselinestretch{.9}
%\bibliographystyle{elsarticle-num} 
\bibliographystyle{IEEEtran}
\bibliography{references}

%\bibliography{reid_bib,EE,EarlyFusion,BibLegacy,BibReckNNGraphCC,BibHypergraph,CNN}
%*******************************************************************************************************

\end{document}